\pdfoutput=1
\documentclass{article}

\PassOptionsToPackage{sort&compress}{natbib}
\usepackage{iclr2027_conference,times}

\usepackage{amsmath,amsfonts,bm}

\def\eqref#1{equation~\ref{#1}}

\def\1{\bm{1}}

\DeclareMathAlphabet{\mathsfit}{\encodingdefault}{\sfdefault}{m}{sl}
\SetMathAlphabet{\mathsfit}{bold}{\encodingdefault}{\sfdefault}{bx}{n}

\iclrfinalcopy %

\usepackage[T1]{fontenc}
\usepackage{amsmath}
\usepackage{amsfonts}
\usepackage{amssymb}
\usepackage{graphicx}
\usepackage[hidelinks]{hyperref}  %
\usepackage{url}  %
\usepackage{booktabs}
\usepackage{subcaption}    %
\usepackage{longtable}
\usepackage{pifont}  %

\usepackage[capitalize]{cleveref}

\crefname{section}{Sec.}{Secs.}
\Crefname{section}{Section}{Sections}

\crefname{table}{Tab.}{Tabs.}      %
\Crefname{table}{Table}{Tables}    %

\crefname{figure}{Fig.}{Figs.}     %
\Crefname{figure}{Figure}{Figures} %

\crefname{appendix}{App.}{Apps.}
\Crefname{appendix}{Appendix}{Appendices}
\usepackage{array}          %
\usepackage{xcolor}
\usepackage{colortbl}       %
\usepackage{wrapfig}

\definecolor{kellygreen}{rgb}{0.3, 0.73, 0.09}
\definecolor{alizarin}{rgb}{0.82, 0.1, 0.26}
\definecolor{ZXY}{rgb}{0.21, 0.49, 0.74}  %
\definecolor{tabhead}{HTML}{DCE7F2}   %
\definecolor{tabrule}{HTML}{E4EEF8}   %
\definecolor{tabjudge}{HTML}{EDE6F5}  %

\usepackage{comment}

\usepackage{csquotes}

\usepackage{fancyvrb}
\fvset{fontsize=\footnotesize, frame=single, framesep=4pt}

\usepackage[most]{tcolorbox}
\definecolor{cardframe}{HTML}{5A87B0}  %
\definecolor{cardback}{HTML}{F4F8FC}
\newtcolorbox{rulecard}[2]{enhanced, width=0.86\linewidth, center,
  colback=cardback, colframe=cardframe, boxrule=0.6pt, arc=2mm,
  left=2.5mm, right=2.5mm, top=1.5mm, bottom=1.5mm, before skip=8pt, after skip=8pt,
  fonttitle=\small\bfseries, coltitle=white, title={#1},
  fontupper=\small, before upper={\setlength{\parskip}{3pt}{\color{gray}\footnotesize #2}\par}}

\usepackage{fontawesome5}

\title{Skill Training with \\ Corruption and Reconstruction Loop}

\author{
  Mo Li\textsuperscript{1,2,4} \hspace{0.9em}
  Zixin Yin\textsuperscript{3,4} \hspace{0.9em}
  Qihao Wu\textsuperscript{4} \hspace{0.9em} \\
  \textbf{Ting Cao}\textsuperscript{1} \hspace{0.9em}
  \textbf{Yunxin Liu}\textsuperscript{1} \hspace{0.9em}
  \textbf{Heung-Yeung Shum}\textsuperscript{1} \\[0.35em]
  {\small
   \textsuperscript{1}Tsinghua University \hspace{0.8em}
   \textsuperscript{2}Shanghai AI Laboratory} \\[0.15em]
  {\small
   \textsuperscript{3}The Hong Kong University of Science and Technology \hspace{0.8em}
   \textsuperscript{4}Xiaobing.AI}
}

\begin{document}

\maketitle

\begin{abstract}

Large Language Models (LLMs) often struggle in highly specialized domains. Rather than parameter-level adaptation of LLMs which is costly and difficult to interpret, external skills (often defined as text files) have been recently proposed to augment LLMs for specialized domains. However, such skills rely on costly active human annotations or passive summarization of high-quality examples.
In this paper, we propose a self-supervised approach for agent self-evolution that learns domain-specific skills directly from existing high-quality human artifacts, without additional human annotations or external rewards. 
Inspired by diffusion models, our approach follows a forward--loss--backward process to reconstruct human artifacts by iteratively learning the agent's external skill library rather than updating its model parameters. 
Experiments on short-drama screenwriting demonstrate that our approach enables agents to autonomously extract generalizable writing skills from human-authored scripts and substantially improve domain-specific generation quality. 
Our approach provides a scalable paradigm for agents to continuously learn many kinds of complex skills from existing high-quality human artifacts.

\vspace{0.6em}
\centerline{\normalsize \href{https://github.com/skilltraining-project/skill-training}{\faGithub\ Code} \hspace{2em} \href{https://skilltraining-project.github.io}{\faGlobe\ Project Page}}
\end{abstract}

\section{Introduction}\label{sec:introduction}

Large Language Models (LLMs) have demonstrated remarkable capabilities in general instruction following and broad knowledge representation~\citep{foundation-models, instructgpt, gpt3}.
However, they often underperform human experts in highly specialized domains, such as repository-specific software development~\citep{li2026learningcommitgeneratingorganic} and complex creative writing~\citep{one-sentence-one-drama}.
A natural approach is to fine-tune these models, yet this is increasingly impractical: the strongest models are closed-source, and fine-tuning frontier-scale open-source models is prohibitively expensive even with parameter-efficient methods~\citep{lora}.
Fine-tuning also risks catastrophic forgetting~\citep{llm-catastrophic-forgetting}.
Whatever is learned through fine-tuning is also tied to a single checkpoint and does not transfer when the base model is upgraded.
Even when fine-tuning is feasible, whatever is learned gets buried inside the model weights, where practitioners can neither inspect what was acquired nor correct what was learned wrong~\citep{lima, urial, reward-overoptimization, reward-hacking-monitoring}.
These limitations motivate an alternative paradigm: rather than modifying model parameters, can we enable agents to acquire specialized capabilities through externalized skills?

Recent works explore this direction by augmenting LLM agents with external \enquote{skills}~\citep{voyager, expel, anthropic-agent-skills}, which are portable across different models and human-auditable, allowing agents to behave like a human expert in a plug-and-play way.
However, the effectiveness of this paradigm ultimately depends on the quality and coverage of the skills it can acquire, while existing approaches rely heavily on human experts to manually create domain-specific skills.
This reliance creates a significant bottleneck: manually creating skills is costly and labor-intensive, while domain experts may be reluctant to formalize their implicit knowledge for fear of having it \enquote{distilled}.
Consequently, enabling agents to autonomously acquire high-quality domain-specific skills, without additional human effort, remains a fundamental challenge.

To automate skill acquisition without additional human feedback or external rewards, we propose a self-supervised framework inspired by diffusion models~\citep{sohldickstein2015deepunsupervisedlearningusing, ho2020denoisingdiffusionprobabilisticmodels}.
We treat high-quality human artifacts as the \enquote{clean} state, summarize them into an intermediate \enquote{noisy} state, and ask the agent to reconstruct the original artifacts from these summaries~\citep{bart-denoising-seq2seq}.
This reconstruction task provides a natural learning signal for open-ended domains, where verifiable rewards are typically unavailable.
More importantly, comparing the agent's reconstruction with the human original reveals where the agent falls short and provides targeted textual feedback for self-reflection.
The agent then distills these differences into generalizable skills rather than simply memorizing individual artifacts, allowing existing human-authored artifacts to serve as supervision for autonomous skill acquisition.

As the agent accumulates skills from more human-authored artifacts, the skill library, or memory, grows over time, raising two challenges: how to organize an expanding collection of skills and how to update them based on new learning signals.
To organize the growing skill library, we adopt a Mixture-of-Experts (MoE) architecture~\citep{skill-based-mixture-of-experts}, which separates skills from each other so that the agent activates only a relevant subset for each task.
To update these skills, we formulate skill evolution as a textual analogue of parameter optimization~\citep{textgrad, protegi}: in the forward stage, the agent uses relevant skills to generate an artifact; in the loss stage, it compares the generated artifact with the human original to identify what is missing or incorrect; and in the backward stage, it revises the skills most relevant to the observed shortcomings.

These components enable agents to acquire and continually refine domain-specific skills from existing human-authored artifacts, without additional human feedback or external rewards. Our contributions are as follows:

\begin{itemize}
    \item We propose a self-supervised framework that trains an external skill library rather than model weights, which helps automatically extract skills from human artifacts.
    \item We validate the framework on short-drama screenwriting, an open-ended domain where verifiable rewards are unavailable. We also work with professional practitioners to design ten event-based metrics that evaluate specific writing behaviors in generated scripts.
    \item We conduct diverse experiments on skill training and analyze its behavior as a training process. We release our code at \href{https://github.com/skilltraining-project/skill-training}{github.com/skilltraining-project/skill-training} to accelerate community research on skill training.
\end{itemize}

\section{Related Work}\label{sec:related}

\paragraph{Skill Acquisition for LLM Agents.}
Instead of fine-tuning model weights, a growing body of work equips LLM agents with external skill libraries that guide their behavior in specialized domains~\citep{voyager, expel, coala, anthropic-agent-skills}.
There are several ways to build such a library.
The most straightforward is to have human experts write the skills directly \citep{co-writing-screenplays-and-theatre-scripts-with-language-models-an-evaluation-by-industry-professionals}, but this is costly and experts may resist formalizing their implicit knowledge.
Automated alternatives extract skills from agent trajectories, either by storing complete trajectories for retrieval \citep{memento, coala} or by distilling them into abstract skill documents \citep{voyager, expel, reasoningbank}.
But both require successful trajectories, which are easy to obtain when verifiable rewards exist (e.g., coding or math)~\citep{skillopt, skillopt-lite}, but it is hard to judge whether a trajectory is successful at all in open-ended domains like screenwriting~\citep{openmeva-story-generation-metrics, large-language-models-cannot-self-correct-reasoning-yet}.
A third approach that works for open-ended domains bypasses trajectories and directly summarizes high-quality examples into rules~\citep{honovich2022instructioninductionexamplesnatural}.
We follow this path, but instead of passively summarizing, we draw on the corruption-and-reconstruction philosophy of diffusion models~\citep{sohldickstein2015deepunsupervisedlearningusing, ho2020denoisingdiffusionprobabilisticmodels} to construct a learning signal: we first corrupt the human artifact into a summary, ask the agent to reconstruct the original~\citep{bart-denoising-seq2seq}, and learn from the gap between the reconstruction and the human version.

\paragraph{Skill Organization and Updating.}
Once a reliable learning signal is obtained, the agent must update its skill library efficiently.
Early memory mechanisms append experiences to a flat linear stream \citep{generative-agents, reflexion}, which makes specific memories hard to locate and update as the pool grows.
To solve the locating problem, recent work organizes skill documents into hierarchical structures \citep{skillgrad}. We adopt a similar idea with a Mixture-of-Experts (MoE) routing mechanism \citep{skill-based-mixture-of-experts}, so that accumulated skills stay localized and different cards cover different aspects of writing.
To solve the updating problem, recent work maps backpropagation into the textual domain, using language models to generate natural language feedback for credit assignment across text nodes \citep{textgrad, protegi}.
We follow this direction: our backward agent routes textual losses to the relevant skills while model parameters stay frozen, shifting gradient updates from the parameter space to the external skill library.

\section{Method}\label{sec:method}

\begin{figure}[!t]
\centering
\includegraphics[width=\linewidth]{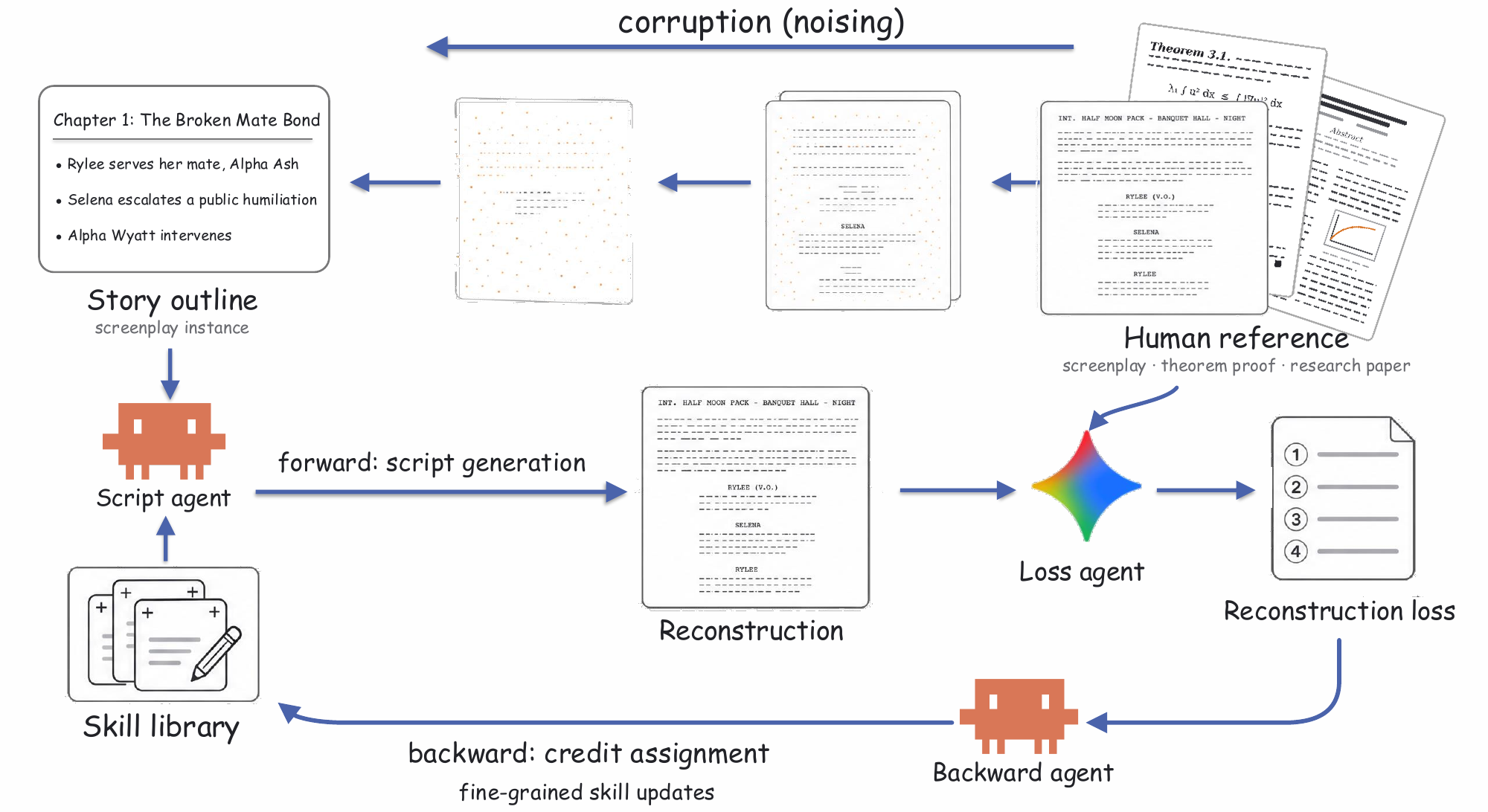}
\caption{Overview of the framework. A human screenplay is corrupted into a condensed story outline (corruption). The script agent reconstructs the screenplay from the outline with the help of the skill library (reconstruction); a loss agent contrasts the reconstruction against the human original to produce a textual loss; a backward agent routes this loss into the skill library as fine-grained updates. The reconstruction loop iterates to continue to improve reconstruction quality (loop). The same corruption-reconstruction-loop framework applies to any domain beyond screenplay that offers human reference artifacts like theorem proving or research writing.}
\label{fig:framework}
\end{figure}

Our framework treats a skill library as the trainable parameters of the agent, optimizing it through a corruption-and-reconstruction loop as illustrated in \Cref{fig:framework}.

\subsection{Corruption: Building Self-Supervised Pairs}\label{sec:method:corruption}

We start from professional, production-quality short-drama screenplays.
For each episode, a corruption step compresses the script into a condensed story outline, stripping away dialogue, physical actions, and pacing.
Similar to the forward process in diffusion models, this destroys craft-level details while preserving the story skeleton.
The resulting (outline, original screenplay) pairs serve as self-supervised training examples.
The choice of the corruption operator can define the skill to be learned: stripping craft forces the agent to recover craft.
This same design generalizes to other domains, such as erasing mathematical proofs while keeping theorems to train proof construction.

\subsection{Forward Script Generation}\label{sec:method:forward}

Given the noised outline, the script agent reconstructs the full episode in Hollywood format.
Before writing, the agent retrieves and reads relevant rule cards from the skill library.
The system logs the complete generation trajectory alongside a read trace recording which cards were activated.

\subsection{Semantic Loss}\label{sec:method:loss}

The backward update requires a loss signal that captures the semantic gap between the reconstruction and the human original.
In creative writing, token-level matching does not work, as two excellent scripts may share no exact surface text.
We therefore use a loss agent that reads the two scripts side by side and produces a dense semantic loss report.
\textbf{We do not prescribe which dimensions to compare.} The agent is simply given both scripts and asked to identify differences, allowing it to discover relevant aspects of screenwriting craft on its own.

\subsection{Backward Credit Assignment}\label{sec:method:backward}

Given the loss report and the read trace, the next step is to assign credit to individual rule cards and update them accordingly.
We use a backward agent as the optimizer to carry out this assignment.
Updates follow four operations: strengthening cards that provided correct guidance, narrowing the scope of cards that misled the agent, creating new cards for recurring unhandled mistakes, and merging redundant rules so that each card stays focused on a distinct aspect of writing.

\begin{figure}[!t]
\centering
\includegraphics[width=\linewidth]{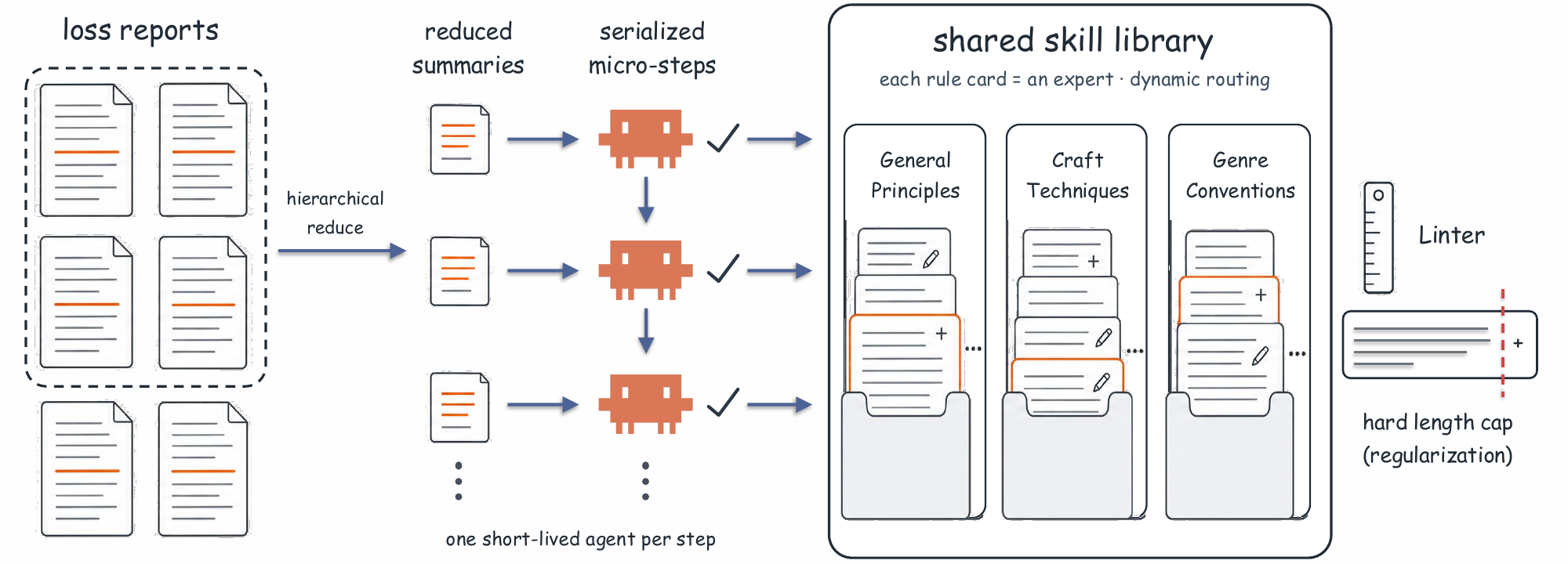}
\caption{Backward credit assignment with memory regularization and MoE routing. Loss reports are reduced into summaries (left), processed by serialized micro-steps (center), and routed into expert folders with a linter enforcing token limits (right).}
\label{fig:memory_arch}
\end{figure}

\subsection{Memory Regularization and MoE Routing}\label{sec:method:memory}

As the skill library grows through training, two problems arise: how to keep the library organized so that the agent can find the right skills, and how to prevent individual cards from overfitting to specific novels.
To address organization, we structure the skill library using the Mixture-of-Experts (MoE) topology shown in \Cref{fig:memory_arch}, routing rule cards into specialized folders (general principles, craft techniques, and genre conventions). \Cref{sec:appendix:library_content} shows example cards.
To address overfitting, we initially allowed unconstrained growth, but early experiments showed the library aggressively memorizing case details.
After manual inspection, we found the longest card had grown to 2{,}500 lines, reading more like incident logs of each novel than general rules.
We therefore introduce a rule linter that acts as a hard regularizer.
Each card body is capped at 1{,}200 tokens and each description at 200 tokens, checked at every micro-step commit.
This hard capacity forces the backward agent to distill abstract rules and discard instance-level details.

\subsection{Scaling with Data-Parallel Micro-Steps}\label{sec:method:scale}

Each training step begins with a single forward pass: the agent writes the next five episodes of every training novel in parallel against the current snapshot of the skill library, and the loss agent scores each reconstruction against the human original.
This produces a batch of loss reports, all reflecting the same pool state.
Processing this entire batch at once causes signal dilution: the backward agent produces vague edits when overwhelmed by context.
We therefore split the backward pass into micro-steps.
Loss reports are first grouped and reduced into compact summaries of cross-novel commonalities.
Each micro-step then consumes one reduced summary, applies a small focused set of edits to the shared skill library, and terminates before the next begins.
Because all loss reports originate from the same forward snapshot but micro-steps update the pool sequentially, later steps operate on a pool that has already changed.
We address this by explicitly prompting the backward agent with this time gap, so it validates each edit against the current pool before applying it.

\section{Experiments}\label{sec:experiments}

We evaluate our framework on short-drama screenwriting, where the agent generates full screenplays from story outlines, and test whether skills accumulated through training improve the agent's screenwriting on entirely unseen stories.

\subsection{Experimental Setup}\label{sec:exp:setup}

\begin{wrapfigure}{r}{0.3\linewidth}
\vspace{-1.3em}
\centering
\includegraphics[width=\linewidth]{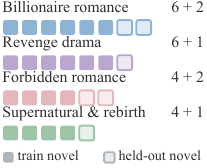}
\caption{Genre distribution of the 20 training and 6 held-out novels.}
\label{fig:genre}
\vspace{-2.5em}
\end{wrapfigure}
\paragraph{Training configuration.} Our training corpus consists of twenty proprietary short-drama novels written by professional screenwriters, all of which have been produced into high-performing series on TikTok.
As shown in \Cref{fig:genre}, the novels cover four popular short-drama genres, led by billionaire romance and revenge drama, followed by forbidden romance and supernatural or rebirth stories.
Each novel contains 49 to 70 episodes averaging roughly 2{,}000 characters per episode in Hollywood format.
Each training run spans three epochs with the same hyperparameters and a 1{,}200-token-per-card length limit.
We use Claude Sonnet 4.6 for the main experiments and also test the training loop with Kimi K2.6~\citep{kimiteam2026kimik25visualagentic} and Claude Opus 4.8.

\paragraph{Test set and baselines.} To evaluate whether the learned skills generalize to unseen stories, we hold out six novels with a total of 385 episodes that never appear in any training run.
For each held-out novel, we generate every episode under four configurations with the same execution context budget.
\textbf{Base} uses an empty skill library. \textbf{+Few-shot} provides human-written scripts as in-context examples. \textbf{+Distill} learns rules directly from human originals, using the same training novels and number of micro-steps as \textbf{+Ours}, which learns through the reconstruction loop.

\begin{table}[!t]
\centering
\caption{Definitions of the ten evaluation metrics. Line-level style metrics are counted by a rule-based parser. Episode-level structure metrics are judged by an LLM with quoted evidence.}
\label{tab:metrics}
\small
\begin{tabular*}{\linewidth}{@{\extracolsep{\fill}} p{0.26\linewidth} p{0.68\linewidth}@{}}
\toprule
Metric & Definition \\
\midrule
\rowcolor{tabrule}[0pt][0pt]\multicolumn{2}{@{}l@{}}{\textit{Line-level style}} \\
Longest dialogue volley $\sim$ & Longest run of speaker turns between action lines, scenes, or transitions \\
Dialogue line length $\sim$ & Average words per dialogue line across the episode \\
Voice-over cues $\sim$ & Count of voice-over cues per episode \\
Unfilmable action lines $\downarrow$ & Psychological action lines that a camera cannot capture \\
Stimulus words $\uparrow$ & Count of dramatic-stimulus words per episode \\
\midrule
\rowcolor{tabjudge}[0pt][0pt]\multicolumn{2}{@{}l@{}}{\textit{Episode-level structure}} \\
Plot-conflict density $\uparrow$ & Distinct confrontations between opposing parties, per thousand words \\
Hard-cut ending $\uparrow$ & The episode stops mid-action rather than after the outcome lands \\
Cold open $\uparrow$ & Conflict starts from the very first beat \\
Prop activation $\uparrow$ & Key props get physically used, not just gazed at \\
Ending twist $\sim$ & The final scene overturns the direction the episode had set up, rather than merely escalating it \\
\bottomrule
\end{tabular*}
\end{table}

\paragraph{Evaluation.} Evaluating open-ended screenwriting is difficult.
Existing screenplay benchmarks target feature-film aesthetics or generic script continuation~\citep{cmlbench}, which differ sharply from what vertical short drama requires.
A common alternative is to have an LLM judge score each script on a 1-to-5 scale, but we find such judgments unreliable because the judge's own variance is large enough that the same script can receive a 2 on one run and a 4 on the next.
We therefore work with professional short-drama practitioners to design the ten proxy metrics listed in \Cref{tab:metrics}, each anchored to a concrete, verifiable event rather than a subjective impression. These metrics measure specific writing behaviors and do not directly capture overall script quality, which we assess separately through human evaluation in \Cref{sec:exp:human}.
The metrics fall into two groups by the level of the script they describe.
Five line-level style metrics count features of individual lines and are extracted by a deterministic parser.
Five episode-level structure metrics describe how the whole episode is built and are judged by an LLM with quoted evidence. The judge counts plot conflicts, which we normalize by script length, and answers yes, no, or na for the four binary checks.

We report episode-level means for counts and densities, and yes-rates among valid judgments for binary checks.
Of the ten metrics, six have a clear preferred direction ($\uparrow$ or $\downarrow$), drawn from both groups. The other four ($\sim$) track properties like line length and voice-over frequency, where professional scripts themselves vary widely. We report them alongside human values to show how each method changes the writing style.
As a supplementary comparison, we report paired mean absolute error for counts and densities, $d_g = \frac{1}{n}\sum_{i=1}^{n}|x_{g,i}-h_i|$, where $x_{g,i}$ and $h_i$ are the method's and human's values for the same episode. This measures agreement with the original, rather than determining which method performs best.
Significance tests compare each method's raw values or binary verdicts with Base on the same episodes, using 10{,}000 paired bootstrap resamples as detailed in \Cref{sec:appendix:stats}.
We use Gemini 3.1 Pro Preview as the primary judge and cross-check with Claude Opus 4.8, finding 94\% agreement across all verdicts, confirming that the event-based metrics produce stable results regardless of which model serves as the judge.

\subsection{Does the Reconstruction Loop Help?}\label{sec:exp:results}

\begin{table}[!t]
\centering
\caption{Method comparison. Columns Base through Human report mean counts or densities and binary yes-rates. Bold marks the highest $\uparrow$ or lowest $\downarrow$ value among methods; $\sim$ denotes descriptive metrics without a preferred direction. In the value columns, $^{*}$ marks differences from Base whose 95\% paired bootstrap intervals exclude zero. The auxiliary $d$ columns report paired MAE against human originals for counts and densities, or method-minus-human rate gaps in percentage points.}
\label{tab:main}
\small
\resizebox{\linewidth}{!}{%
\begin{tabular}{lrrrr>{\columncolor{black!6}}rrrrr}
\toprule
Metric & Base & +Few-shot & +Distill & +Ours & Human & $d_{\text{Base}}$ & $d_{\text{+FS}}$ & $d_{\text{+Distill}}$ & $d_{\text{+Ours}}$ \\
\midrule
\rowcolor{tabrule}\multicolumn{10}{l}{\textit{Line-level style}} \\
Longest dialogue volley (turns/ep)~$\sim$ & 4.5 & 4.6 & 4.9$^{*}$ & 4.2$^{*}$ & 3.6 & 1.86 & 1.94 & 2.05 & 1.72 \\
Dialogue line length (words)~$\sim$ & 5.6 & 10.2$^{*}$ & 7.1$^{*}$ & 7.3$^{*}$ & 11.9 & 6.53 & 3.54 & 5.47 & 4.99 \\
Voice-over cues (per ep)~$\sim$ & 0.35 & 0.37 & 2.56$^{*}$ & 0.52$^{*}$ & 0.22 & 0.48 & 0.51 & 2.36 & 0.58 \\
Unfilmable action lines (per ep)~$\downarrow$ & 1.2 & 1.5$^{*}$ & 1.5$^{*}$ & \textbf{0.2}$^{*}$ & 0.4 & 1.09 & 1.36 & 1.38 & 0.46 \\
Stimulus words (per ep)~$\uparrow$ & 3.3 & 2.9$^{*}$ & 3.3 & \textbf{6.3}$^{*}$ & 9.2 & 6.54 & 6.64 & 6.31 & 5.01 \\
\midrule
\rowcolor{tabjudge}\multicolumn{10}{l}{\textit{Episode-level structure}} \\
Plot conflicts (per 1k words)~$\uparrow$ & 2.10 & 2.00 & 1.81$^{*}$ & \textbf{3.69}$^{*}$ & 3.47 & 2.17 & 2.29 & 2.39 & 2.06 \\
Hard-cut ending rate~$\uparrow$ & 34\% & 34\% & 32\% & \textbf{67\%}$^{*}$ & 62\% & $-28.3$ & $-28.3$ & $-30.4$ & $+4.7$ \\
Cold-open rate~$\uparrow$ & 52\% & 52\% & 52\% & \textbf{74\%}$^{*}$ & 59\% & $-7.3$ & $-7.0$ & $-6.5$ & $+15.1$ \\
Prop-activation rate~$\uparrow$ & 46\% & 45\% & 48\% & \textbf{58\%}$^{*}$ & 57\% & $-11.4$ & $-12.4$ & $-9.0$ & $+0.8$ \\
Ending-twist rate~$\sim$ & 33\% & 34\% & 32\% & 31\% & 31\% & $+1.6$ & $+2.3$ & $+0.3$ & $-0.8$ \\
\bottomrule
\end{tabular}}

\end{table}

We compare four configurations in \Cref{tab:main}.
To ensure a fair comparison, the generation agent operates under the same context budget across all settings.
The \textbf{+Few-shot} baseline includes 24 diverse human-written episodes (spanning various genres and story stages) in the prompt.
While this approach successfully transfers surface statistics, raising dialogue length from 5.6 to 10.2 words against a human 11.9, it fails to significantly improve any of the four binary craft checks compared to the Base configuration.
Finished scripts show the agent what professional writing looks like, but not how its own writing fails.
Distilling rules directly from the 20 training novels (\textbf{+Distill}) goes a step further.
It accumulates 131 rule cards totaling over 114{,}000 tokens, yet episode-level structure does not improve: hard-cut endings show no gain, at 32\% against the base 34\%.
Both baselines share a fundamental blind spot: professional scripts contain no amateur flaws.
Consequently, neither human examples nor distilled rules can teach the agent what it specifically does wrong.

The reconstruction loop (\textbf{+Ours}) fills this gap.
Because the agent generates scripts from story outlines and a loss agent compares these drafts against human originals, the training signal captures both what the human does right and what the agent does wrong.
Although the +Ours library learns only 23 rule cards from the novels, a fraction of the 131 cards generated by +Distill, it significantly improves all six directional metrics over Base.
Hard-cut endings rise from 34\% to 67\%, past the human 62\%, and unfilmable action lines drop from 1.2 to 0.2 per episode.
Prop activation, cold opens, stimulus words, and conflict density all improve significantly as well.
The stylistic metrics ($\sim$) show how each method changes the writing beyond these targeted techniques.
Dialogue volleys shorten from 4.5 turns in Base to 4.2 with +Ours, as the agent breaks up extended exchanges with action beats and scene transitions.
+Distill moves in the opposite direction, stretching volleys to 4.9 turns and pushing voice-over use to 2.56 cues per episode, seven times the base rate of 0.35, while +Ours stays at 0.52.
Ending-twist rates stay roughly flat as expected, since the outlines already determine most of the twists.

\subsection{Do the Learned Skills Transfer Across Models?}\label{sec:exp:xmodel}

\begin{figure}[!t]
\centering
\includegraphics[width=\linewidth]{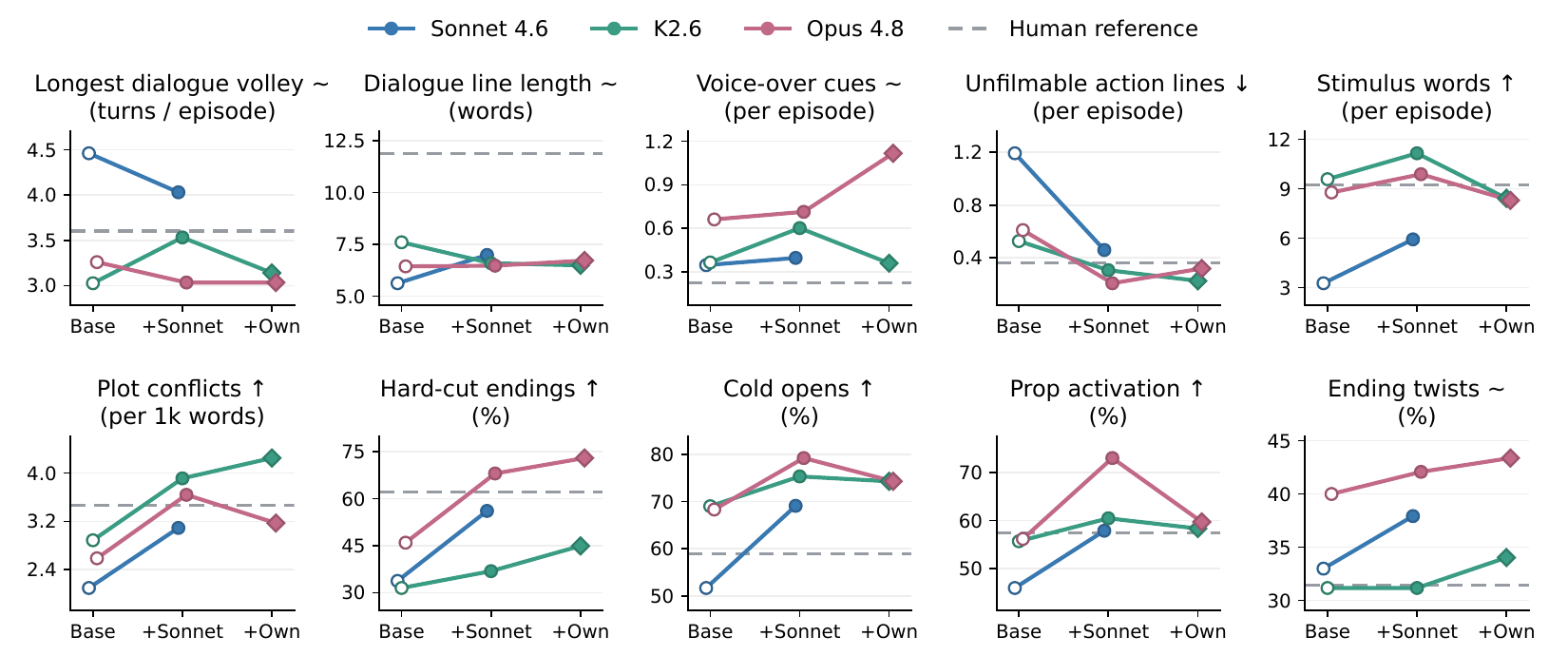}
\caption{Cross-model evaluation with empty, transferred Sonnet, and own-model skill libraries. For Opus, the Base-to-+Sonnet segments run roughly parallel to Sonnet's own on most metrics, showing that the transferred library brings Opus gains similar to those it brings Sonnet, while divergence with own-model libraries reflects the error-driven nature of skill acquisition. All libraries are trained on five novels. Arrows mark preferred directions; $\sim$ marks descriptive metrics.}
\label{fig:xmodel}
\end{figure}

The reconstruction loop learns from the generating model's own errors. We next ask whether other models can learn through the same loop and whether they can directly reuse Sonnet's learned skills.
We evaluate Kimi K2.6 and Claude Opus 4.8 with their own libraries and with Sonnet's transferred library.
\Cref{fig:xmodel} shows the results, and \Cref{tab:xmodel_full} in \Cref{sec:appendix:xmodel} lists them in full.
Both models successfully improve using Sonnet's library, showing that distilled skills transfer across models.
Counterintuitively, Opus scores significantly higher on five of six directional metrics with Sonnet's library than with its own.
\Cref{tab:paired_ci} in \Cref{sec:appendix:stats} reports these paired comparisons.

This gap suggests that the learning loop is largely error-driven: models mainly distill rules for the mistakes they actually make.
Base Opus is highly capable (e.g., natively generating 8.8 stimulus words per episode), so it makes few of the foundational errors that the loop corrects.
Consequently, it formulates far fewer basic quantitative rules.
Conversely, the weaker Sonnet frequently fails at basic techniques, forcing it to compile a rigorous library of hard constraints.
When Opus executes Sonnet's library, it pairs its superior baseline with Sonnet's strict rules, yielding peak performance.
The sole exception is hard-cut endings. Opus is relatively weak here without a library, at 46\%, so its own library targets this metric and raises it to 73\%.
These results point to a practical paradigm: using weaker, cheaper models to mine foundational rules for stronger models to execute.
\Cref{sec:appendix:xmodel_library} compares the composition of the two libraries.

\subsection{Do Human Readers Prefer Scripts Written with Learned Skills?}\label{sec:exp:human}

We understand that the proxy metrics can only measure specific writing behaviors but may not reflect a reader's overall impression of a script.
We therefore conduct a blind preference study with five readers, three external short-drama practitioners and two paper authors.
In the trained-library versus empty-library comparison, the trained version wins 49 of 75 judgments (65\%).
Notably, the professional screenwriter prefers the trained version over the human original in 9 of 15 pairs.
The skills learned through the reconstruction loop improve not only the proxy metrics but also the overall quality perceived by human readers.
\Cref{sec:appendix:human} reports all three comparisons, per-reader breakdowns, confidence intervals, and the bootstrap methodology.

\section{Analysis}\label{sec:analysis}

The experiments above show that the reconstruction loop produces useful skills.
We now ask how the learned skill library changes as the amount of training data grows and as training proceeds, and whether the agent can apply these skills in a production pipeline unseen during training.

\subsection{How Does Corpus Size Affect the Learned Skills?}\label{sec:exp:scaling}

\begin{figure}[!t]
\centering
\includegraphics[width=\linewidth]{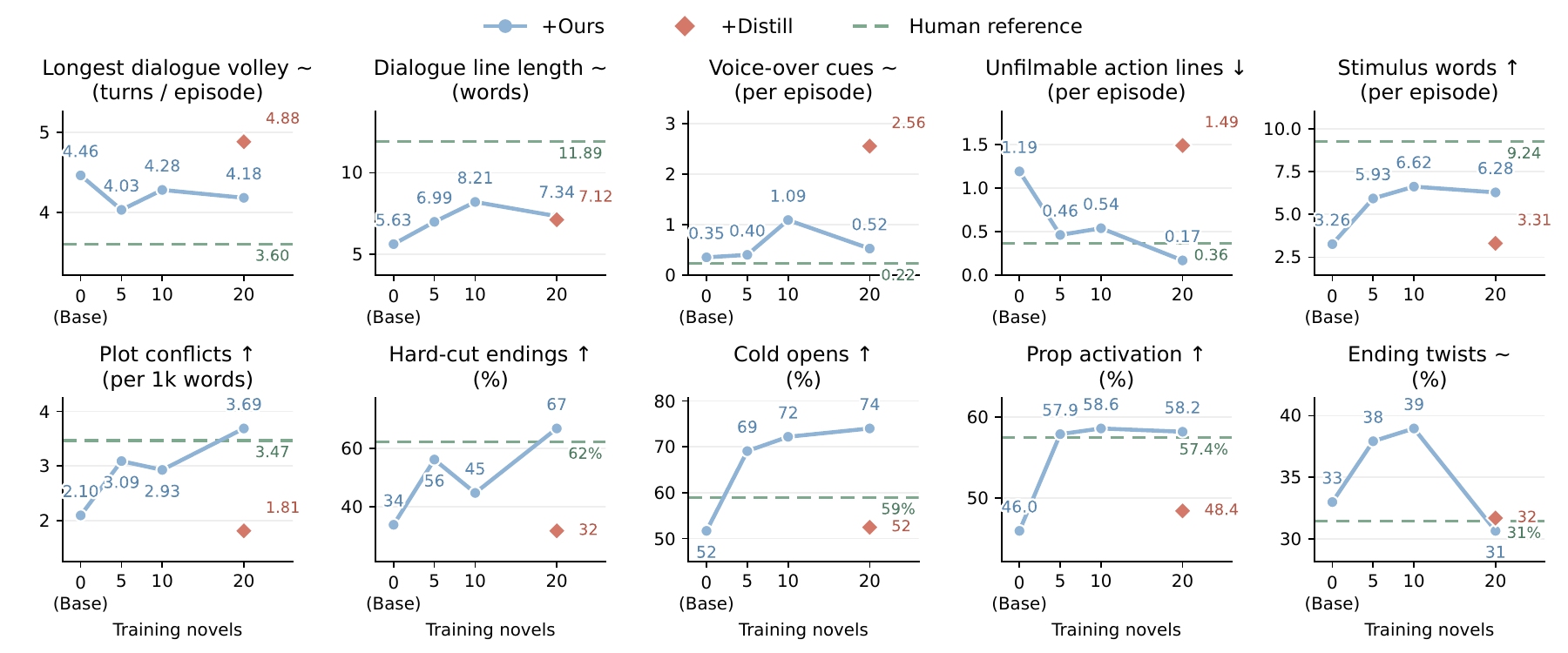}
\caption{Scaling the training corpus from zero to twenty novels. Panels report mean counts or densities and binary yes-rates. Blue circles trace reconstruction training, red diamonds mark +Distill trained on twenty novels, and dashed lines mark human references. Arrows indicate preferred directions; $\sim$ marks descriptive metrics.}
\label{fig:scaling}
\end{figure}

The twenty-novel results raise the question of how much training data is needed for these gains.
We compare libraries trained on 5, 10, and 20 novels on the same held-out test set in \Cref{fig:scaling}.
\Cref{tab:scaling_full} in \Cref{sec:appendix:scaling} reports the complete results, including paired MAE.
With just five novels, hard-cut endings rise from 34\% to 56\%, cold opens from 52\% to 69\%, and prop activation from 46\% to 58\% relative to Base.
With only a quarter of the training novels used by +Distill, the five-novel library already achieves higher rates on all three measures.

Expanding the training corpus from five to twenty novels improves all six directional metrics numerically, although prop activation changes little.
Hard-cut endings rise from 56\% to 67\% and cold opens from 69\% to 74\%.
Unfilmable action lines fall from 0.46 to 0.17 per episode, while plot conflicts increase from 3.09 to 3.69 per thousand words.

At the intermediate scale of ten novels, however, hard-cut endings fall to 45\% before recovering to 67\% at twenty novels. Voice-over use also changes, rising from 0.40 to 1.09 cues per episode and then falling to 0.52.
To understand this pattern, we inspect the learned rules.
The ten-novel library encourages voice-overs when dialogue conceals a character's feelings and allows an initial reaction before the episode cuts. The twenty-novel library narrows voice-overs to a few uses, forbids them at episode endings, and emphasizes cutting immediately after key actions or revelations.
These differences align with the observed rise and fall in voice-over use and the opposite pattern in hard-cut endings.
Because skills are stored as explicit text cards, we can inspect the writing instructions associated with these performance changes.
\Cref{sec:appendix:library_content} compares the content of the three libraries rule by rule and shows what each larger corpus adds.
Overall, five novels are enough to learn useful screenwriting skills, while scaling to twenty further improves the targeted writing techniques.

\subsection{How Does the Agent's Behavior Change as It Trains?}\label{sec:exp:dynamics}

\begin{figure}[!t]
\centering
\includegraphics[width=\linewidth]{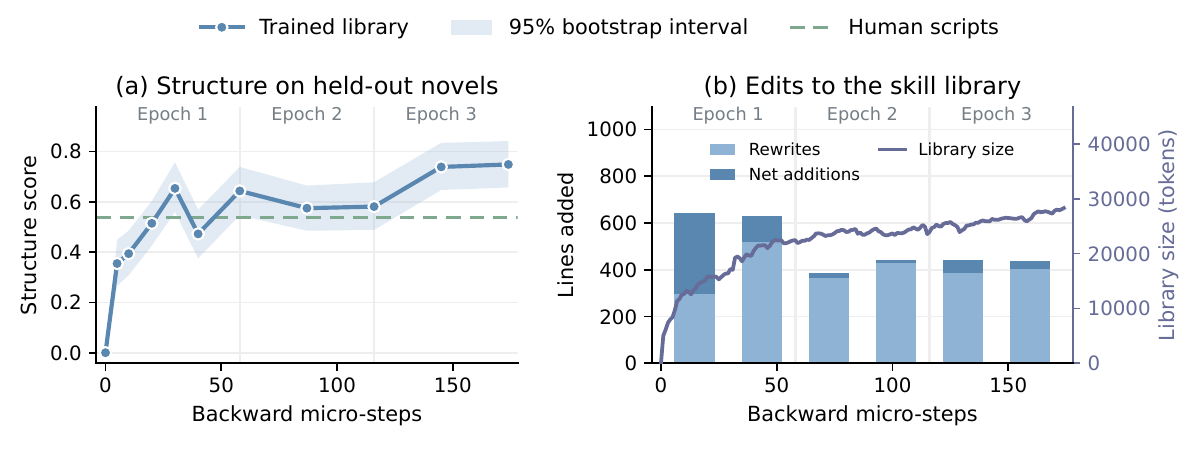}
\caption{Twenty-novel training run. (a) Structure score of skill library snapshots on the first 30 episodes of each held-out novel. The band is a 95\% bootstrap interval over episodes, and the dashed line scores human scripts on the same scale. (b) Lines added to the skill library in each half epoch of 29 micro-steps, split into net additions and rewrites that replace deleted lines. The line shows the library size in tokens.}
\label{fig:training_dynamics}
\end{figure}

Besides more training data, longer training is another way to scale. We therefore ask whether the skill library keeps improving as training goes on.
We fix the corpus at twenty novels and evaluate snapshots of the library from its 174 backward micro-steps on the held-out novels.
We combine the four directional episode-level structure metrics into one structure score (details in \Cref{sec:appendix:dynamics}).
The score rises quickly in the first epoch, passes the human reference between micro-steps 20 and 30, and levels off near its highest value of 0.75 in the third epoch.
As \Cref{fig:dynamics_metrics} in \Cref{sec:appendix:dynamics} shows, line-level style settles even faster, within five micro-steps.

\Cref{fig:training_dynamics}b shows a matching pattern in the library itself.
In the first 29 micro-steps, 54\% of the added lines are net additions that grow the library.
In later half epochs, this share stays between 3\% and 18\%, so the backward agent spends the rest of training revising the rules it has already written.

\subsection{Can the Learned Skills Generalize to a Third-Party Production Pipeline?}\label{sec:exp:case}

The preceding analyses track how skills develop within our training framework.
However, in real production scenarios, the generation environment may not match our training setup, and third-party pipelines may require outputs in a tabular storyboard format rather than Hollywood-format screenplays.
In \Cref{sec:appendix:case}, we deploy a Sonnet 4.6 skill library trained on twenty novels into such a pipeline, generating five episodes of an unseen drama series in a format the agent never encountered during training.
We find the learned skills adapt effectively to the new format: the agent enforces hard-cut endings, translates internal monologue into filmable actions, and demonstrates in-context self-correction.
However, possessing a rule does not guarantee its application: in one episode, the agent does not activate the ending rule and produces a static fade-out.
These results suggest that the skills our framework acquires generalize well across frameworks, but reliable skill activation remains an important direction for future improvement.

\section{Conclusion and Future Work}\label{sec:conclusion}

We introduced a self-supervised framework that shifts agent optimization from parameter fine-tuning to the iterative refinement of an external, human-readable skill library. By leveraging a corruption-and-reconstruction loop on existing high-quality human artifacts, agents can autonomously acquire and generalize specialized domain skills without requiring newly collected human annotations or external reward modeling. 

However, the effectiveness of this paradigm depends heavily on two critical factors, which also point out our next steps. First, the textual credit assignment chain breaks if the base model reads a retrieved rule but fails to execute it, rendering any updates to the skill library ineffective. Second, a rule is only as useful as the retrieval mechanism that accurately surfaces it for a specific scene. Enhancing the robustness of both instruction-following and retrieval is our next step. Furthermore, by incorporating human feedback during real-world deployment, this training loop could naturally extend into a continual learning framework, enabling the agent to incrementally self-evolve its skills.

\ificlrfinal
\section*{Author Contributions}

Mo Li proposed the main idea and conducted the main experiments. Zixin Yin contributed to the method design, annotated data, and engaged in weekly discussions. Qihao Wu coordinated screenplay resources and organized discussions with professional screenwriters. Ting Cao and Yunxin Liu participated in discussions and provided feedback. Heung-Yeung Shum provided computing resources, engaged in weekly discussions, gave extensive feedback, and revised the paper.
\fi

\bibliographystyle{iclr2027_conference}
\bibliography{references}

\begin{thebibliography}{34}
\providecommand{\natexlab}[1]{#1}
\providecommand{\url}[1]{\texttt{#1}}
\expandafter\ifx\csname urlstyle\endcsname\relax
  \providecommand{\doi}[1]{doi: #1}\else
  \providecommand{\doi}{doi: \begingroup \urlstyle{rm}\Url}\fi

\bibitem[Baker et~al.(2025)Baker, Huizinga, Gao, Dou, Guan, Madry, Zaremba,
  Pachocki, and Farhi]{reward-hacking-monitoring}
Bowen Baker, Joost Huizinga, Leo Gao, Zehao Dou, Melody~Y. Guan, Aleksander
  Madry, Wojciech Zaremba, Jakub Pachocki, and David Farhi.
\newblock Monitoring reasoning models for misbehavior and the risks of
  promoting obfuscation, 2025.
\newblock URL \url{https://arxiv.org/abs/2503.11926}.

\bibitem[Bommasani et~al.(2022)Bommasani, Hudson, Adeli, Altman, Arora, von
  Arx, Bernstein, Bohg, Bosselut, Brunskill, Brynjolfsson, Buch, Card,
  Castellon, Chatterji, Chen, Creel, Davis, Demszky, Donahue, Doumbouya,
  Durmus, Ermon, Etchemendy, Ethayarajh, Fei-Fei, Finn, Gale, Gillespie, Goel,
  Goodman, Grossman, Guha, Hashimoto, Henderson, Hewitt, Ho, Hong, Hsu, Huang,
  Icard, Jain, Jurafsky, Kalluri, Karamcheti, Keeling, Khani, Khattab, Koh,
  Krass, Krishna, Kuditipudi, Kumar, Ladhak, Lee, Lee, Leskovec, Levent, Li,
  Li, Ma, Malik, Manning, Mirchandani, Mitchell, Munyikwa, Nair, Narayan,
  Narayanan, Newman, Nie, Niebles, Nilforoshan, Nyarko, Ogut, Orr,
  Papadimitriou, Park, Piech, Portelance, Potts, Raghunathan, Reich, Ren, Rong,
  Roohani, Ruiz, Ryan, Ré, Sadigh, Sagawa, Santhanam, Shih, Srinivasan,
  Tamkin, Taori, Thomas, Tramèr, Wang, Wang, Wu, Wu, Wu, Xie, Yasunaga, You,
  Zaharia, Zhang, Zhang, Zhang, Zhang, Zheng, Zhou, and
  Liang]{foundation-models}
Rishi Bommasani, Drew~A. Hudson, Ehsan Adeli, Russ Altman, Simran Arora, Sydney
  von Arx, Michael~S. Bernstein, Jeannette Bohg, Antoine Bosselut, Emma
  Brunskill, Erik Brynjolfsson, Shyamal Buch, Dallas Card, Rodrigo Castellon,
  Niladri Chatterji, Annie Chen, Kathleen Creel, Jared~Quincy Davis, Dora
  Demszky, Chris Donahue, Moussa Doumbouya, Esin Durmus, Stefano Ermon, John
  Etchemendy, Kawin Ethayarajh, Li~Fei-Fei, Chelsea Finn, Trevor Gale, Lauren
  Gillespie, Karan Goel, Noah Goodman, Shelby Grossman, Neel Guha, Tatsunori
  Hashimoto, Peter Henderson, John Hewitt, Daniel~E. Ho, Jenny Hong, Kyle Hsu,
  Jing Huang, Thomas Icard, Saahil Jain, Dan Jurafsky, Pratyusha Kalluri,
  Siddharth Karamcheti, Geoff Keeling, Fereshte Khani, Omar Khattab, Pang~Wei
  Koh, Mark Krass, Ranjay Krishna, Rohith Kuditipudi, Ananya Kumar, Faisal
  Ladhak, Mina Lee, Tony Lee, Jure Leskovec, Isabelle Levent, Xiang~Lisa Li,
  Xuechen Li, Tengyu Ma, Ali Malik, Christopher~D. Manning, Suvir Mirchandani,
  Eric Mitchell, Zanele Munyikwa, Suraj Nair, Avanika Narayan, Deepak
  Narayanan, Ben Newman, Allen Nie, Juan~Carlos Niebles, Hamed Nilforoshan,
  Julian Nyarko, Giray Ogut, Laurel Orr, Isabel Papadimitriou, Joon~Sung Park,
  Chris Piech, Eva Portelance, Christopher Potts, Aditi Raghunathan, Rob Reich,
  Hongyu Ren, Frieda Rong, Yusuf Roohani, Camilo Ruiz, Jack Ryan, Christopher
  Ré, Dorsa Sadigh, Shiori Sagawa, Keshav Santhanam, Andy Shih, Krishnan
  Srinivasan, Alex Tamkin, Rohan Taori, Armin~W. Thomas, Florian Tramèr,
  Rose~E. Wang, William Wang, Bohan Wu, Jiajun Wu, Yuhuai Wu, Sang~Michael Xie,
  Michihiro Yasunaga, Jiaxuan You, Matei Zaharia, Michael Zhang, Tianyi Zhang,
  Xikun Zhang, Yuhui Zhang, Lucia Zheng, Kaitlyn Zhou, and Percy Liang.
\newblock On the opportunities and risks of foundation models, 2022.
\newblock URL \url{https://arxiv.org/abs/2108.07258}.

\bibitem[Brown et~al.(2020)Brown, Mann, Ryder, Subbiah, Kaplan, Dhariwal,
  Neelakantan, Shyam, Sastry, Askell, Agarwal, Herbert-Voss, Krueger, Henighan,
  Child, Ramesh, Ziegler, Wu, Winter, Hesse, Chen, Sigler, Litwin, Gray, Chess,
  Clark, Berner, McCandlish, Radford, Sutskever, and Amodei]{gpt3}
Tom~B. Brown, Benjamin Mann, Nick Ryder, Melanie Subbiah, Jared Kaplan,
  Prafulla Dhariwal, Arvind Neelakantan, Pranav Shyam, Girish Sastry, Amanda
  Askell, Sandhini Agarwal, Ariel Herbert-Voss, Gretchen Krueger, Tom Henighan,
  Rewon Child, Aditya Ramesh, Daniel~M. Ziegler, Jeffrey Wu, Clemens Winter,
  Christopher Hesse, Mark Chen, Eric Sigler, Mateusz Litwin, Scott Gray,
  Benjamin Chess, Jack Clark, Christopher Berner, Sam McCandlish, Alec Radford,
  Ilya Sutskever, and Dario Amodei.
\newblock Language models are few-shot learners, 2020.
\newblock URL \url{https://arxiv.org/abs/2005.14165}.

\bibitem[Chen et~al.(2026)Chen, Yun, Stengel-Eskin, Chen, and
  Bansal]{skill-based-mixture-of-experts}
Justin Chih-Yao Chen, Sukwon Yun, Elias Stengel-Eskin, Tianlong Chen, and Mohit
  Bansal.
\newblock Skill-based mixture-of-experts: Adaptive routing for heterogeneous
  reasoning via inferred skills, 2026.
\newblock URL \url{https://arxiv.org/abs/2503.05641}.

\bibitem[Gao et~al.(2022)Gao, Schulman, and Hilton]{reward-overoptimization}
Leo Gao, John Schulman, and Jacob Hilton.
\newblock Scaling laws for reward model overoptimization, 2022.
\newblock URL \url{https://arxiv.org/abs/2210.10760}.

\bibitem[Guan et~al.(2021)Guan, Zhang, Feng, Liu, Ding, Mao, Fan, and
  Huang]{openmeva-story-generation-metrics}
Jian Guan, Zhexin Zhang, Zhuoer Feng, Zitao Liu, Wenbiao Ding, Xiaoxi Mao,
  Changjie Fan, and Minlie Huang.
\newblock Openmeva: A benchmark for evaluating open-ended story generation
  metrics, 2021.
\newblock URL \url{https://arxiv.org/abs/2105.08920}.

\bibitem[Ho et~al.(2020)Ho, Jain, and
  Abbeel]{ho2020denoisingdiffusionprobabilisticmodels}
Jonathan Ho, Ajay Jain, and Pieter Abbeel.
\newblock Denoising diffusion probabilistic models, 2020.
\newblock URL \url{https://arxiv.org/abs/2006.11239}.

\bibitem[Honovich et~al.(2022)Honovich, Shaham, Bowman, and
  Levy]{honovich2022instructioninductionexamplesnatural}
Or~Honovich, Uri Shaham, Samuel~R. Bowman, and Omer Levy.
\newblock Instruction induction: From few examples to natural language task
  descriptions, 2022.
\newblock URL \url{https://arxiv.org/abs/2205.10782}.

\bibitem[Hu et~al.(2021)Hu, Shen, Wallis, Allen-Zhu, Li, Wang, Wang, and
  Chen]{lora}
Edward~J. Hu, Yelong Shen, Phillip Wallis, Zeyuan Allen-Zhu, Yuanzhi Li, Shean
  Wang, Lu~Wang, and Weizhu Chen.
\newblock Lora: Low-rank adaptation of large language models, 2021.
\newblock URL \url{https://arxiv.org/abs/2106.09685}.

\bibitem[Huang et~al.(2024)Huang, Chen, Mishra, Zheng, Yu, Song, and
  Zhou]{large-language-models-cannot-self-correct-reasoning-yet}
Jie Huang, Xinyun Chen, Swaroop Mishra, Huaixiu~Steven Zheng, Adams~Wei Yu,
  Xinying Song, and Denny Zhou.
\newblock Large language models cannot self-correct reasoning yet, 2024.
\newblock URL \url{https://arxiv.org/abs/2310.01798}.

\bibitem[Lewis et~al.(2019)Lewis, Liu, Goyal, Ghazvininejad, Mohamed, Levy,
  Stoyanov, and Zettlemoyer]{bart-denoising-seq2seq}
Mike Lewis, Yinhan Liu, Naman Goyal, Marjan Ghazvininejad, Abdelrahman Mohamed,
  Omer Levy, Ves Stoyanov, and Luke Zettlemoyer.
\newblock Bart: Denoising sequence-to-sequence pre-training for natural
  language generation, translation, and comprehension, 2019.
\newblock URL \url{https://arxiv.org/abs/1910.13461}.

\bibitem[Li et~al.(2026)Li, Xu, Tan, Cao, and
  Liu]{li2026learningcommitgeneratingorganic}
Mo~Li, L.~H. Xu, Qitai Tan, Ting Cao, and Yunxin Liu.
\newblock Learning to commit: Generating organic pull requests via online
  repository memory, 2026.
\newblock URL \url{https://arxiv.org/abs/2603.26664}.

\bibitem[Lin et~al.(2023)Lin, Ravichander, Lu, Dziri, Sclar, Chandu,
  Bhagavatula, and Choi]{urial}
Bill~Yuchen Lin, Abhilasha Ravichander, Ximing Lu, Nouha Dziri, Melanie Sclar,
  Khyathi Chandu, Chandra Bhagavatula, and Yejin Choi.
\newblock The unlocking spell on base llms: Rethinking alignment via in-context
  learning, 2023.
\newblock URL \url{https://arxiv.org/abs/2312.01552}.

\bibitem[Luo et~al.(2025)Luo, Yang, Meng, Li, Zhou, and
  Zhang]{llm-catastrophic-forgetting}
Yun Luo, Zhen Yang, Fandong Meng, Yafu Li, Jie Zhou, and Yue Zhang.
\newblock An empirical study of catastrophic forgetting in large language
  models during continual fine-tuning, 2025.
\newblock URL \url{https://arxiv.org/abs/2308.08747}.

\bibitem[Mirowski et~al.(2022)Mirowski, Mathewson, Pittman, and
  Evans]{co-writing-screenplays-and-theatre-scripts-with-language-models-an-evaluation-by-industry-professionals}
Piotr Mirowski, Kory~W. Mathewson, Jaylen Pittman, and Richard Evans.
\newblock Co-writing screenplays and theatre scripts with language models: An
  evaluation by industry professionals, 2022.
\newblock URL \url{https://arxiv.org/abs/2209.14958}.

\bibitem[Ouyang et~al.(2022)Ouyang, Wu, Jiang, Almeida, Wainwright, Mishkin,
  Zhang, Agarwal, Slama, Ray, Schulman, Hilton, Kelton, Miller, Simens, Askell,
  Welinder, Christiano, Leike, and Lowe]{instructgpt}
Long Ouyang, Jeff Wu, Xu~Jiang, Diogo Almeida, Carroll~L. Wainwright, Pamela
  Mishkin, Chong Zhang, Sandhini Agarwal, Katarina Slama, Alex Ray, John
  Schulman, Jacob Hilton, Fraser Kelton, Luke Miller, Maddie Simens, Amanda
  Askell, Peter Welinder, Paul Christiano, Jan Leike, and Ryan Lowe.
\newblock Training language models to follow instructions with human feedback,
  2022.
\newblock URL \url{https://arxiv.org/abs/2203.02155}.

\bibitem[Ouyang et~al.(2026)Ouyang, Yan, Hsu, Chen, Jiang, Wang, Han, Le,
  Daruki, Tang, Tirumalashetty, Lee, Rofouei, Lin, Han, Lee, and
  Pfister]{reasoningbank}
Siru Ouyang, Jun Yan, I-Hung Hsu, Yanfei Chen, Ke~Jiang, Zifeng Wang, Rujun
  Han, Long~T. Le, Samira Daruki, Xiangru Tang, Vishy Tirumalashetty, George
  Lee, Mahsan Rofouei, Hangfei Lin, Jiawei Han, Chen-Yu Lee, and Tomas Pfister.
\newblock Reasoningbank: Scaling agent self-evolving with reasoning memory,
  2026.
\newblock URL \url{https://arxiv.org/abs/2509.25140}.

\bibitem[Park et~al.(2023)Park, O'Brien, Cai, Morris, Liang, and
  Bernstein]{generative-agents}
Joon~Sung Park, Joseph~C. O'Brien, Carrie~J. Cai, Meredith~Ringel Morris, Percy
  Liang, and Michael~S. Bernstein.
\newblock Generative agents: Interactive simulacra of human behavior, 2023.
\newblock URL \url{https://arxiv.org/abs/2304.03442}.

\bibitem[Pryzant et~al.(2023)Pryzant, Iter, Li, Lee, Zhu, and Zeng]{protegi}
Reid Pryzant, Dan Iter, Jerry Li, Yin Lee, Chenguang Zhu, and Michael Zeng.
\newblock Automatic prompt optimization with {\enquote{gradient descent}} and
  beam search.
\newblock In Houda Bouamor, Juan Pino, and Kalika Bali (eds.),
  \emph{Proceedings of the 2023 Conference on Empirical Methods in Natural
  Language Processing}, pp.\  7957--7968, Singapore, December 2023. Association
  for Computational Linguistics.
\newblock \doi{10.18653/v1/2023.emnlp-main.494}.
\newblock URL \url{https://aclanthology.org/2023.emnlp-main.494/}.

\bibitem[Shen et~al.(2026)Shen, Li, and Zhang]{skillopt-lite}
Yifei Shen, Bo~Li, and Xinjie Zhang.
\newblock Skillopt-lite: Better and faster agent self-evolution via one line of
  vibe, 2026.
\newblock URL \url{https://arxiv.org/abs/2607.03451}.

\bibitem[Shi et~al.(2026)Shi, Yan, Huang, Chen, Zhang, He, Yeo, and
  Li]{one-sentence-one-drama}
Yufei Shi, Weilong Yan, Naixuan Huang, Yucheng Chen, Chenyu Zhang, Tao He,
  Si~Yong Yeo, and Ming Li.
\newblock One sentence, one drama: Personalized short-form drama generation via
  multi-agent systems, 2026.
\newblock URL \url{https://arxiv.org/abs/2605.22144}.

\bibitem[Shinn et~al.(2023)Shinn, Cassano, Berman, Gopinath, Narasimhan, and
  Yao]{reflexion}
Noah Shinn, Federico Cassano, Edward Berman, Ashwin Gopinath, Karthik
  Narasimhan, and Shunyu Yao.
\newblock Reflexion: Language agents with verbal reinforcement learning, 2023.
\newblock URL \url{https://arxiv.org/abs/2303.11366}.

\bibitem[Sohl-Dickstein et~al.(2015)Sohl-Dickstein, Weiss, Maheswaranathan, and
  Ganguli]{sohldickstein2015deepunsupervisedlearningusing}
Jascha Sohl-Dickstein, Eric~A. Weiss, Niru Maheswaranathan, and Surya Ganguli.
\newblock Deep unsupervised learning using nonequilibrium thermodynamics, 2015.
\newblock URL \url{https://arxiv.org/abs/1503.03585}.

\bibitem[Sumers et~al.(2024)Sumers, Yao, Narasimhan, and Griffiths]{coala}
Theodore~R. Sumers, Shunyu Yao, Karthik Narasimhan, and Thomas~L. Griffiths.
\newblock Cognitive architectures for language agents, 2024.
\newblock URL \url{https://arxiv.org/abs/2309.02427}.

\bibitem[Team et~al.(2026)Team, Bai, Bai, Bao, Cai, Cao, Chai, Charles, Che,
  Chen, Chen, Chen, Chen, Chen, Chen, Chen, Chen, Chen, Chen, Chen, Chen, Chen,
  Chen, Chen, Chen, Chen, Chen, Chen, Chen, Cheng, Cheng, Chu, Cui, Deng, Diao,
  Ding, Dong, Dong, Dong, Dong, Du, Du, Du, Du, Du, Fan, Fang, Feng, Feng, Fu,
  Fu, Gao, Gao, Ge, Geng, Gong, Gong, Gongque, Gu, Gu, Gu, Guan, Guan, Guo,
  Hao, He, He, He, He, He, He, Hong, Hu, Hu, Hu, Hu, Huang, Huang, Huang,
  Huang, Jia, Jiang, Jiang, Jin, Jing, Lai, Li, Li, Li, Li, Li, Li, Li, Li, Li,
  Li, Li, Li, Li, Li, Li, Li, Li, Li, Li, Li, Li, Li, Li, Li, Liao, Lin, Lin,
  Lin, Lin, Lin, Liu, Liu, Liu, Liu, Liu, Liu, Liu, Liu, Liu, Liu, Liu, Liu,
  Liu, Liu, Liu, Liu, Liu, Lu, Lu, Lu, Luo, Luo, Luo, Luo, Ma, Mao, Mei, Men,
  Meng, Meng, Miao, Ni, Ouyang, Pan, Pang, Qian, Qin, Qin, Qiu, Qu, Shang,
  Shao, Shen, Shen, Shi, Shi, Shi, Song, Song, Song, Song, Su, Su, Su, Sui,
  Sun, Sun, Sun, Sung, Tai, Tang, Tang, Tang, Tang, Tao, Teng, Tian, Tian,
  Wang, Wang, Wang, Wang, Wang, Wang, Wang, Wang, Wang, Wang, Wang, Wang, Wang,
  Wang, Wang, Wang, Wang, Wang, Wang, Wang, Wang, Wang, Wang, Wang, Wang, Wang,
  Wang, Wang, Wang, Wang, Wang, Wang, Wang, Wang, Wang, Wang, Wang, Wang, Wang,
  Wei, Wei, Wen, Wen, Wu, Wu, Wu, Wu, Wu, Wu, Wu, Wu, Wu, Xiao, Xie, Xie, Xie,
  Xing, Xu, Xu, Xu, Xu, Xu, Xu, Xu, Xu, Xu, Xu, Xu, Xu, Xu, Xu, Xu, Yan, Yan,
  Yang, Yang, Yang, Yang, Yang, Yang, Yang, Yang, Yang, Yang, Yang, Yang, Yang,
  Yang, Yao, Ye, Ye, Ye, Ye, Yebo, Yin, Yu, Yu, Yu, Yu, Yuan, Yuan, Yuan, Yue,
  Zeng, Zha, Zhan, Zhang, Zhang, Zhang, Zhang, Zhang, Zhang, Zhang, Zhang,
  Zhang, Zhang, Zhang, Zhang, Zhang, Zhang, Zhang, Zhang, Zhang, Zhang, Zhang,
  Zhao, Zhao, Zhao, Zhao, Zhao, Zhao, Zhao, Zhao, Zheng, Zheng, Zheng, Zheng,
  Zhong, Zhong, Zhong, Zhou, Zhou, Zhou, Zhou, Zhu, Zhu, Zhu, Zhu, Zhu, Zhuang,
  Zhuang, Zou, and Zu]{kimiteam2026kimik25visualagentic}
Kimi Team, Tongtong Bai, Yifan Bai, Yiping Bao, S.~H. Cai, Yuan Cao, Ziwei
  Chai, Y.~Charles, H.~S. Che, Cheng Chen, Guanduo Chen, Huarong Chen, Jia
  Chen, Jianlong Chen, Jun Chen, Kefan Chen, Liang Chen, Ruijue Chen, Xinhao
  Chen, Yanru Chen, Yanxu Chen, Yicun Chen, Yimin Chen, Yingjiang Chen, Yuankun
  Chen, Yujie Chen, Yutian Chen, Zhirong Chen, Ziwei Chen, Dazhi Cheng, Yean
  Cheng, Minghan Chu, Jialei Cui, Jiaqi Deng, Muxi Diao, Hao Ding, Mengfan
  Dong, Mengnan Dong, Yuxin Dong, Yuhao Dong, Angang Du, Chenzhuang Du, Dikang
  Du, Lingxiao Du, Yulun Du, Yu~Fan, Shengjun Fang, Qiulin Feng, Yichen Feng,
  Garimugai Fu, Kelin Fu, Hongcheng Gao, Tong Gao, Yuyao Ge, Shangyi Geng,
  Chengyang Gong, Xiaochen Gong, Zhuoma Gongque, Qizheng Gu, Xinran Gu, Yicheng
  Gu, Longyu Guan, Shuhao Guan, Yuanying Guo, Xiaoru Hao, Dailan He, Tianhong
  He, Weiran He, Wenyang He, Yibo He, Yunjia He, Chao Hong, Hao Hu, Jiaxi Hu,
  Yangyang Hu, Zhenxing Hu, Ke~Huang, Ruiyuan Huang, Weixiao Huang, Zhiqi
  Huang, Chaobo Jia, Tao Jiang, Zhejun Jiang, Xinyi Jin, Yu~Jing, Guokun Lai,
  Aidi Li, C.~Li, Cheng Li, Fang Li, Guanghe Li, Guanyu Li, Haitao Li, Haoyang
  Li, Jia Li, Jingwei Li, Junxiong Li, Lincan Li, Mo~Li, Weihong Li, Wentao Li,
  Xinhang Li, Xinhao Li, Yang Li, Yanhao Li, Yiwei Li, Yuxiao Li, Zhaowei Li,
  Zhaoxi Li, Zheming Li, Weilong Liao, Jiawei Lin, Xiaohan Lin, Yibo Lin,
  Zhishan Lin, Zichao Lin, Cheng Liu, Chenyu Liu, Hongzhang Liu, Liang Liu,
  Shaowei Liu, Shudong Liu, Shuran Liu, Tianwei Liu, Tianyu Liu, Weizhou Liu,
  Xiangyan Liu, Yangyang Liu, Yanming Liu, Yibo Liu, Yuanxin Liu, Zhengying
  Liu, Zhongnuo Liu, Enzhe Lu, Haoyu Lu, Zhiyuan Lu, G.~Luo, Junyu Luo, Tongxu
  Luo, Yashuo Luo, Long Ma, Shaoguang Mao, Yuan Mei, Xin Men, Fanqing Meng,
  Zhiyong Meng, Yibo Miao, Minqing Ni, Kun Ouyang, Siyuan Pan, Bo~Pang, Yuchao
  Qian, Ruoyu Qin, Zeyu Qin, Jiezhong Qiu, Bowen Qu, Zeyu Shang, Youbo Shao,
  Tianxiao Shen, Zhennan Shen, Juanfeng Shi, Lidong Shi, Shengyuan Shi, Feifan
  Song, Pengwei Song, Tianhui Song, Xiaoxi Song, Hongjin Su, Jianlin Su,
  Zhaochen Su, Lin Sui, Jinsong Sun, Junyao Sun, Tongyu Sun, Flood Sung,
  Yunpeng Tai, Chuning Tang, Heyi Tang, Xiaojuan Tang, Zhengyang Tang, Jiawen
  Tao, Shiyuan Teng, Chaoran Tian, Pengfei Tian, Bowen Wang, Chensi Wang,
  Chuang Wang, Congcong Wang, Dingkun Wang, Dinglu Wang, Dongliang Wang, Feng
  Wang, Hailong Wang, Haiming Wang, Hao Wang, Hengzhi Wang, Huaqing Wang, Hui
  Wang, Jiahao Wang, Jinhong Wang, Jiuzheng Wang, Kaixin Wang, Linian Wang,
  Qibin Wang, Shengjie Wang, Shuyi Wang, Si~Wang, Wei Wang, Xiaochen Wang,
  Xinyuan Wang, Yao Wang, Yejie Wang, Yipu Wang, Yiqin Wang, Yucheng Wang,
  Yuzhi Wang, Zhaoji Wang, Zhaowei Wang, Zhengtao Wang, Zhexu Wang, Zifan Wang,
  Zihan Wang, Zizhe Wang, Chu Wei, Ming Wei, Chuan Wen, Zichen Wen, Chengjie
  Wu, Haoning Wu, Junyan Wu, Rucong Wu, Wenhao Wu, Yuefeng Wu, Yuhao Wu, Yuxin
  Wu, Zijian Wu, Chenjun Xiao, Jin Xie, Xiaotong Xie, Yuchong Xie, Bowei Xing,
  Boyu Xu, Jianfan Xu, Jing Xu, Jinjing Xu, L.~H. Xu, Lin Xu, Suting Xu, Weixin
  Xu, Xinbo Xu, Xinran Xu, Yangchuan Xu, Yichang Xu, Yuemeng Xu, Zelai Xu,
  Ziyao Xu, Junjie Yan, Yuzi Yan, Guangyao Yang, Hao Yang, Junwei Yang, Kai
  Yang, Ningyuan Yang, Xiaofei Yang, Xinlong Yang, Xinyu Yang, Ying Yang,
  Yi~Yang, Yi~Yang, Zhen Yang, Zhilin Yang, Zonghan Yang, Haotian Yao, Dan Ye,
  Haoran Ye, Wenjie Ye, Zhuorui Ye, Peng Yebo, Bohong Yin, Chengzhen Yu,
  Longhui Yu, Tao Yu, Tianxiang Yu, Enming Yuan, Mengjie Yuan, Xiaokun Yuan,
  Yang Yue, Weihao Zeng, Dunyuan Zha, Haobing Zhan, Dehao Zhang, Hao Zhang, Jin
  Zhang, Puqi Zhang, Qiao Zhang, Rui Zhang, Xiaobin Zhang, Xiaoyun Zhang,
  Y.~Zhang, Yadong Zhang, Yangkun Zhang, Yichi Zhang, Yizhi Zhang, Yongting
  Zhang, Yu~Zhang, Yushun Zhang, Yutao Zhang, Yutong Zhang, Zheng Zhang,
  Chenguang Zhao, Feifan Zhao, Jinxiang Zhao, Shuai Zhao, Xiangyu Zhao, Xuanle
  Zhao, Yikai Zhao, Zijia Zhao, Huabin Zheng, Ruihan Zheng, Shaojie Zheng,
  Tengyang Zheng, Junfeng Zhong, Longguang Zhong, Weiming Zhong, M.~Zhou,
  Runjie Zhou, Xinyu Zhou, Zaida Zhou, Jinguo Zhu, Liya Zhu, Xinhao Zhu, Yuxuan
  Zhu, Zhen Zhu, Jingze Zhuang, Weiyu Zhuang, Ying Zou, and Xinxing Zu.
\newblock Kimi k2.5: Visual agentic intelligence, 2026.
\newblock URL \url{https://arxiv.org/abs/2602.02276}.

\bibitem[Wang et~al.(2023)Wang, Xie, Jiang, Mandlekar, Xiao, Zhu, Fan, and
  Anandkumar]{voyager}
Guanzhi Wang, Yuqi Xie, Yunfan Jiang, Ajay Mandlekar, Chaowei Xiao, Yuke Zhu,
  Linxi Fan, and Anima Anandkumar.
\newblock Voyager: An open-ended embodied agent with large language models,
  2023.
\newblock URL \url{https://arxiv.org/abs/2305.16291}.

\bibitem[Wang et~al.(2026)Wang, Lan, Cao, Lin, and Chen]{skillgrad}
Hanyu Wang, Yifan Lan, Bochuan Cao, Lu~Lin, and Jinghui Chen.
\newblock Skillgrad: Optimizing agent skills like gradient descent, 2026.
\newblock URL \url{https://arxiv.org/abs/2605.27760}.

\bibitem[Yang et~al.(2026)Yang, Gong, Huang, Yang, Zhou, Huang, Li, Gao, Dai,
  Liu, Qiu, Yang, Chen, Yang, and Luo]{skillopt}
Yifan Yang, Ziyang Gong, Weiquan Huang, Qihao Yang, Ziwei Zhou, Zisu Huang, Yan
  Li, Xuemei Gao, Qi~Dai, Bei Liu, Kai Qiu, Yuqing Yang, Dongdong Chen, Xue
  Yang, and Chong Luo.
\newblock Skillopt: Executive strategy for self-evolving agent skills, 2026.
\newblock URL \url{https://arxiv.org/abs/2605.23904}.

\bibitem[Yuksekgonul et~al.(2024)Yuksekgonul, Bianchi, Boen, Liu, Huang,
  Guestrin, and Zou]{textgrad}
Mert Yuksekgonul, Federico Bianchi, Joseph Boen, Sheng Liu, Zhi Huang, Carlos
  Guestrin, and James Zou.
\newblock Textgrad: Automatic {\enquote{differentiation}} via text, 2024.
\newblock URL \url{https://arxiv.org/abs/2406.07496}.

\bibitem[Zhang et~al.(2025)Zhang, Lazuka, and Murag]{anthropic-agent-skills}
Barry Zhang, Keith Lazuka, and Mahesh Murag.
\newblock Equipping agents for the real world with agent skills, 2025.
\newblock URL
  \url{https://www.anthropic.com/engineering/equipping-agents-for-the-real-world-with-agent-skills}.
\newblock Anthropic Engineering Blog.

\bibitem[Zhao et~al.(2024)Zhao, Huang, Xu, Lin, Liu, and Huang]{expel}
Andrew Zhao, Daniel Huang, Quentin Xu, Matthieu Lin, Yong-Jin Liu, and Gao
  Huang.
\newblock Expel: Llm agents are experiential learners, 2024.
\newblock URL \url{https://arxiv.org/abs/2308.10144}.

\bibitem[Zheng et~al.(2025)Zheng, Song, Zhou, You, Zhan, Ma, Song, Lim, Chen,
  and Yang]{cmlbench}
Mingzhe Zheng, Dingjie Song, Guanyu Zhou, Jun You, Jiahao Zhan, Xuran Ma,
  Xinyuan Song, Ser-Nam Lim, Qifeng Chen, and Harry Yang.
\newblock Cml-bench: A framework for evaluating and enhancing llm-powered movie
  scripts generation, 2025.
\newblock URL \url{https://arxiv.org/abs/2510.06231}.

\bibitem[Zhou et~al.(2023)Zhou, Liu, Xu, Iyer, Sun, Mao, Ma, Efrat, Yu, Yu,
  Zhang, Ghosh, Lewis, Zettlemoyer, and Levy]{lima}
Chunting Zhou, Pengfei Liu, Puxin Xu, Srini Iyer, Jiao Sun, Yuning Mao, Xuezhe
  Ma, Avia Efrat, Ping Yu, Lili Yu, Susan Zhang, Gargi Ghosh, Mike Lewis, Luke
  Zettlemoyer, and Omer Levy.
\newblock Lima: Less is more for alignment, 2023.
\newblock URL \url{https://arxiv.org/abs/2305.11206}.

\bibitem[Zhou et~al.(2025)Zhou, Chen, Guo, Yan, Lee, Wang, Lee, Zhang, Shao,
  Yang, and Wang]{memento}
Huichi Zhou, Yihang Chen, Siyuan Guo, Xue Yan, Kin~Hei Lee, Zihan Wang, Ka~Yiu
  Lee, Guchun Zhang, Kun Shao, Linyi Yang, and Jun Wang.
\newblock Memento: Fine-tuning llm agents without fine-tuning llms, 2025.
\newblock URL \url{https://arxiv.org/abs/2508.16153}.

\end{thebibliography}

\clearpage
\appendix
\section{Implementation Details}\label{sec:appendix:impl}

This appendix reports the implementation details of our framework that are needed to reproduce the experiments but are not covered in the main text.
We first describe which models run each part of the training loop, then the size of a training step, the format of the skill library and how the agent reads it, the data preparation, the two baselines, and finally the evaluation.

\paragraph{Agents and models.}
The script agent and the backward agent run on the base model under training. In the main experiments this is Claude Sonnet 4.6, called through the Claude Code command-line interface.
Each agent call starts a fresh session without any user-level configuration, so the trajectory depends only on the prompt, the story outline, and the skill library.
For the cross-model experiments, Claude Opus 4.8 runs in the same interface and Kimi K2.6 runs in the Kimi Code command-line interface.
All agents use the default sampling settings of their model.
A forward call is limited to 40 minutes, a loss call to 30 minutes, and a backward micro-step to 90 minutes. A failed agent call is retried up to twice.

The corruption step, the loss agent, and the reduce step that merges loss reports run as direct API calls to Gemini 3.1 Pro Preview, with GPT-5.4 as a fallback when a call fails. In the 20-novel run, 12 of the 687 loss calls fell back to GPT-5.4.
Gemini 3.1 Pro Preview is also the primary evaluation judge.
\textbf{The loss agent never sees the evaluation metrics or the judge prompts.} The five line-level style metrics are counted by a deterministic parser with no model involved, and the judge verdicts are cross-checked with Claude Opus 4.8 (\Cref{sec:exp:setup}).

\paragraph{Training step.}
A single forward call writes five consecutive episodes of one novel. An epoch consists of 14 steps, which covers the longest training novel of 70 episodes, and each run trains for three epochs.
Within a step, forward calls for different novels run concurrently with four workers.
The resulting loss reports are split into groups of four novels in a fixed order, and each group is reduced into one summary that drives one backward micro-step.
The 20-novel run thus performs 174 micro-steps over three epochs and takes about 57 hours of wall-clock time.

\paragraph{Cost.}
At official API rates, the 20-novel main training with Claude Sonnet 4.6 costs approximately \$1,000, comprising \$590 for the script agent (about \$0.86 per forward call), \$340 for the backward agent (\$1.90 per backward micro-step), and \$50 for the loss agent and reduce steps.
The 5-novel and 10-novel runs cost approximately \$230 and \$400, respectively.
Furthermore, encompassing all training, baselines, evaluations, and developmental attempts, the total expenditure for this paper is approximately \$20,000.

\paragraph{Skill library format and retrieval.}
The skill library starts empty, so every card in a trained library is produced by the training loop.
The library is stored as a git repository, and the backward agent commits its own edits at the end of each micro-step.
Each rule card is a Markdown file whose header holds a single \texttt{description} field that states when the card applies. The body contains one section of writing rules.
Besides the token limits in \Cref{sec:method:memory}, the linter caps the whole card file at 50 lines and rejects episode numbers as well as phrasing that records how the model erred and how the human wrote it. Cards therefore state how a scene should be written instead of logging individual cases.
The linter runs before every commit. Re-running it on all 175 commits of the 20-novel run, including the initial one, shows that every commit passes.
In each forward call, the initial prompt sent to the script agent contains the path and description of every card. The script agent opens the full body of whichever cards it judges relevant.
The read trace is extracted from the file-read tool calls and shell commands in the trajectory.
On the held-out novels, the script agent with the 20-novel library reads a median of 9 of the 23 cards per call.

\paragraph{Data and story outlines.}
The source screenplays are English scripts in Hollywood format, delivered as PDF files.
We extract the text with PyMuPDF and recover the indentation of scene headings, character names, parentheticals, and dialogue from the horizontal position of each line. Watermarks, running headers, and page numbers are removed.
The corruption step compresses each episode into five to eight short paragraphs of prose.
It removes all dialogue, small physical actions, camera directions, and scene headings. It keeps character names, relationships, identity reveals, key evidence, and the closing hook.
Across the training novels, an outline has a median length of 195 words per episode, ranging from 112 to 346.

\paragraph{Baselines.}
The +Few-shot group receives 24 human episodes, taken as the first, middle, and last episode of eight training novels chosen to span different genres, for about 14{,}000 tokens in total.
The +Distill group follows the same schedule as +Ours. In each micro-step, the backward agent reads the human originals of five episodes from each of four novels and distills writing rules from them directly, for 174 micro-steps in total. The skill library in this baseline is subject to the same length limit described in \Cref{sec:method:memory}, where each rule card cannot exceed 1{,}200 tokens for the body and 200 tokens for the description, which is enforced at every commit.
The two methods therefore see the same training material in the same order and update their libraries equally often. The only difference is whether the agent learns from human originals directly or from its own reconstruction errors.
On the held-out novels, all four groups share an identical execution context budget. This budget defines a shared maximum length limit of 200{,}000 tokens for the script agent's entire trajectory, which encompasses the prompt, retrieved rule cards or examples, reasoning steps, and the final output.

\paragraph{Evaluation.}
The judge runs at temperature 0 and must support every verdict with a quoted line from the script. Episodes judged \texttt{na} for a check are excluded from its denominator.
For stimulus words, the parser counts words from lists covering violence, emotion, and profanity, and 34 onomatopoeic sound effects.
For unfilmable action lines, it matches psychological verbs and phrases in action lines only, counting each line at most once.
A dialogue volley breaks at any action line, scene heading, or transition.
We validate the parser against 575 hand-labeled screenplay lines, and it agrees with every label.

The training-dynamics analysis in \Cref{sec:exp:dynamics} adds seven intermediate library snapshots, each writing 180 held-out episodes.
Gemini 3.1 Pro Preview declined two of these 1{,}260 episodes on content grounds, one for the four binary checks and one for plot-conflict counting.
GPT-5.6-Sol judged both using the same prompts.

\section{Statistical Significance Tests}\label{sec:appendix:stats}

We use a paired-episode bootstrap with 10{,}000 resamples to compare configurations. Most tests compare a method with its Base model. We also compare each recipient model's own library directly with the transferred Sonnet library. Episodes are matched by novel and episode number, preserving the same story outline across configurations.
For each count or density metric, we match episodes with valid values in both configurations and calculate $\delta_i=x_{Ai}-x_{Bi}$. We resample these paired differences with replacement and take their mean. The test therefore compares the original metric values directly, without subtracting the human reference.
For binary checks, we encode yes as one and no as zero, exclude invalid verdicts, and apply the same procedure. Rate differences are reported in percentage points.
Because a test keeps only episodes that are valid in both configurations, its difference can deviate slightly from the gap between the rates in the main and cross-model tables, which use each group's own valid episodes. The deviation is largest for prop activation, where the judge marks many episodes as having no key prop. The $d$ columns in those tables are computed from unrounded values.
An asterisk marks a difference whose 95\% percentile bootstrap confidence interval excludes zero. In the main and cross-model tables, stars appear beside raw values and compare each method with Base. A significant change is an improvement only if it follows the metric's preferred direction. Metrics marked $\sim$ have no preferred direction and no bold winner.
\Cref{tab:paired_ci} reports each difference, its confidence interval, the number of valid paired episodes $n$, and how many of the six held-out novels have the same direction of change as the pooled result.

\newcommand{\pairedCICaption}{Paired comparisons between configurations. $\Delta$ is the first minus the second, so a positive $\Delta$ means the first has the higher value. Positive differences favor $\uparrow$ metrics and negative differences favor $\downarrow$ metrics; $\sim$ metrics are descriptive. The $n$ column gives valid paired episodes. $^{*}$ marks 95\% bootstrap intervals excluding zero.}
{\small
\begingroup
\setlength{\tabcolsep}{3pt}
\setlength{\LTcapwidth}{\linewidth}
\begin{longtable}{@{}>{\raggedright\arraybackslash}p{\dimexpr.28\linewidth-\tabcolsep\relax}>{\raggedright\arraybackslash}p{\dimexpr.27\linewidth-2\tabcolsep\relax}>{\raggedleft\arraybackslash}p{\dimexpr.12\linewidth-2\tabcolsep\relax}>{\raggedleft\arraybackslash}p{\dimexpr.20\linewidth-2\tabcolsep\relax}>{\raggedleft\arraybackslash}p{\dimexpr.05\linewidth-2\tabcolsep\relax}>{\raggedleft\arraybackslash}p{\dimexpr.08\linewidth-\tabcolsep\relax}@{}}
\caption{\pairedCICaption}\label{tab:paired_ci} \\
\toprule
Comparison & Metric & $\Delta$ & 95\% CI & $n$ & Novels \\
\midrule
\endfirsthead
\multicolumn{6}{l}{\footnotesize\textit{(continued)}} \\
\toprule
Comparison & Metric & $\Delta$ & 95\% CI & $n$ & Novels \\
\midrule
\endhead
\bottomrule
\endlastfoot
Sonnet 4.6: +Ours 5n vs Base & Longest dialogue volley~$\sim$ & -0.43$^{*}$ & [-0.62, -0.24] & 385 & 5/6 \\
 & Dialogue line length~$\sim$ & +1.37$^{*}$ & [+1.07, +1.69] & 383 & 6/6 \\
 & Voice-over cues~$\sim$ & +0.05 & [-0.09, +0.19] & 385 & 4/6 \\
 & Unfilmable action lines~$\downarrow$ & -0.73$^{*}$ & [-0.87, -0.60] & 385 & 6/6 \\
 & Stimulus words~$\uparrow$ & +2.66$^{*}$ & [+2.24, +3.12] & 385 & 6/6 \\
 & Plot conflicts / 1k words~$\uparrow$ & +0.99$^{*}$ & [+0.84, +1.15] & 385 & 6/6 \\
 & Hard-cut ending rate~$\uparrow$ & +22.3\,pt$^{*}$ & [+17.1, +27.3] & 385 & 6/6 \\
 & Cold-open rate~$\uparrow$ & +17.4\,pt$^{*}$ & [+13.0, +22.1] & 385 & 6/6 \\
 & Prop-activation rate~$\uparrow$ & +11.3\,pt$^{*}$ & [+4.7, +17.8] & 213 & 5/6 \\
 & Ending-twist rate~$\sim$ & +4.9\,pt$^{*}$ & [+0.5, +9.4] & 385 & 5/6 \\
\midrule
Sonnet 4.6: +Ours 10n vs Base & Longest dialogue volley~$\sim$ & -0.18 & [-0.39, +0.03] & 385 & 4/6 \\
 & Dialogue line length~$\sim$ & +2.59$^{*}$ & [+2.22, +2.97] & 384 & 5/6 \\
 & Voice-over cues~$\sim$ & +0.74$^{*}$ & [+0.54, +0.94] & 385 & 6/6 \\
 & Unfilmable action lines~$\downarrow$ & -0.65$^{*}$ & [-0.81, -0.50] & 385 & 6/6 \\
 & Stimulus words~$\uparrow$ & +3.36$^{*}$ & [+2.94, +3.79] & 385 & 6/6 \\
 & Plot conflicts / 1k words~$\uparrow$ & +0.83$^{*}$ & [+0.69, +0.97] & 385 & 6/6 \\
 & Hard-cut ending rate~$\uparrow$ & +10.9\,pt$^{*}$ & [+6.0, +15.8] & 385 & 6/6 \\
 & Cold-open rate~$\uparrow$ & +20.5\,pt$^{*}$ & [+16.1, +25.2] & 385 & 6/6 \\
 & Prop-activation rate~$\uparrow$ & +15.2\,pt$^{*}$ & [+8.5, +22.4] & 223 & 5/6 \\
 & Ending-twist rate~$\sim$ & +6.0\,pt$^{*}$ & [+1.3, +10.6] & 385 & 6/6 \\
\midrule
Sonnet 4.6: +Ours 20n vs Base & Longest dialogue volley~$\sim$ & -0.28$^{*}$ & [-0.48, -0.08] & 385 & 5/6 \\
 & Dialogue line length~$\sim$ & +1.72$^{*}$ & [+1.39, +2.05] & 384 & 6/6 \\
 & Voice-over cues~$\sim$ & +0.17$^{*}$ & [+0.03, +0.31] & 385 & 5/6 \\
 & Unfilmable action lines~$\downarrow$ & -1.03$^{*}$ & [-1.16, -0.89] & 385 & 6/6 \\
 & Stimulus words~$\uparrow$ & +3.02$^{*}$ & [+2.59, +3.44] & 385 & 6/6 \\
 & Plot conflicts / 1k words~$\uparrow$ & +1.59$^{*}$ & [+1.40, +1.79] & 385 & 6/6 \\
 & Hard-cut ending rate~$\uparrow$ & +33.0\,pt$^{*}$ & [+27.8, +38.2] & 385 & 6/6 \\
 & Cold-open rate~$\uparrow$ & +22.3\,pt$^{*}$ & [+17.9, +27.0] & 385 & 6/6 \\
 & Prop-activation rate~$\uparrow$ & +14.0\,pt$^{*}$ & [+7.0, +21.4] & 215 & 4/6 \\
 & Ending-twist rate~$\sim$ & -2.3\,pt & [-7.0, +2.1] & 385 & 3/6 \\
\midrule
Sonnet 4.6: +Few-shot vs Base & Longest dialogue volley~$\sim$ & +0.18 & [-0.01, +0.38] & 385 & 5/6 \\
 & Dialogue line length~$\sim$ & +4.53$^{*}$ & [+4.14, +4.94] & 382 & 6/6 \\
 & Voice-over cues~$\sim$ & +0.03 & [-0.18, +0.22] & 385 & 3/6 \\
 & Unfilmable action lines~$\downarrow$ & +0.28$^{*}$ & [+0.10, +0.47] & 385 & 5/6 \\
 & Stimulus words~$\uparrow$ & -0.36$^{*}$ & [-0.66, -0.06] & 385 & 5/6 \\
 & Plot conflicts / 1k words~$\uparrow$ & -0.10 & [-0.23, +0.03] & 385 & 3/6 \\
 & Hard-cut ending rate~$\uparrow$ & +0.0\,pt & [-4.9, +4.9] & 385 & 1/6 \\
 & Cold-open rate~$\uparrow$ & +0.3\,pt & [-4.2, +4.7] & 385 & 3/6 \\
 & Prop-activation rate~$\uparrow$ & -0.9\,pt & [-7.8, +6.0] & 217 & 2/6 \\
 & Ending-twist rate~$\sim$ & +0.8\,pt & [-3.4, +4.9] & 385 & 2/6 \\
\midrule
Sonnet 4.6: 20n library after epoch 1 vs Base & Longest dialogue volley~$\sim$ & -0.16 & [-0.37, +0.05] & 385 & 4/6 \\
 & Dialogue line length~$\sim$ & +2.03$^{*}$ & [+1.60, +2.46] & 384 & 5/6 \\
 & Voice-over cues~$\sim$ & +0.24$^{*}$ & [+0.10, +0.39] & 385 & 4/6 \\
 & Unfilmable action lines~$\downarrow$ & -0.90$^{*}$ & [-1.04, -0.77] & 385 & 6/6 \\
 & Stimulus words~$\uparrow$ & +2.24$^{*}$ & [+1.86, +2.63] & 385 & 6/6 \\
 & Plot conflicts / 1k words~$\uparrow$ & +1.27$^{*}$ & [+1.08, +1.46] & 385 & 6/6 \\
 & Hard-cut ending rate~$\uparrow$ & +28.8\,pt$^{*}$ & [+23.6, +34.0] & 385 & 6/6 \\
 & Cold-open rate~$\uparrow$ & +24.2\,pt$^{*}$ & [+19.5, +28.8] & 385 & 6/6 \\
 & Prop-activation rate~$\uparrow$ & +12.8\,pt$^{*}$ & [+6.0, +19.7] & 218 & 5/6 \\
 & Ending-twist rate~$\sim$ & +3.4\,pt & [-1.0, +7.5] & 385 & 4/6 \\
\midrule
Sonnet 4.6: 20n library after epoch 2 vs Base & Longest dialogue volley~$\sim$ & -0.32$^{*}$ & [-0.52, -0.12] & 385 & 4/6 \\
 & Dialogue line length~$\sim$ & +0.99$^{*}$ & [+0.65, +1.33] & 384 & 4/6 \\
 & Voice-over cues~$\sim$ & -0.10 & [-0.27, +0.05] & 385 & 4/6 \\
 & Unfilmable action lines~$\downarrow$ & -0.91$^{*}$ & [-1.04, -0.78] & 385 & 6/6 \\
 & Stimulus words~$\uparrow$ & +2.37$^{*}$ & [+1.94, +2.82] & 385 & 6/6 \\
 & Plot conflicts / 1k words~$\uparrow$ & +1.48$^{*}$ & [+1.29, +1.68] & 385 & 6/6 \\
 & Hard-cut ending rate~$\uparrow$ & +21.0\,pt$^{*}$ & [+15.8, +26.2] & 385 & 6/6 \\
 & Cold-open rate~$\uparrow$ & +24.4\,pt$^{*}$ & [+19.7, +29.4] & 385 & 6/6 \\
 & Prop-activation rate~$\uparrow$ & +10.1\,pt$^{*}$ & [+3.2, +17.0] & 218 & 5/6 \\
 & Ending-twist rate~$\sim$ & +2.9\,pt & [-1.3, +7.0] & 385 & 4/6 \\
\midrule
K2.6: own library vs Base & Longest dialogue volley~$\sim$ & +0.11 & [-0.05, +0.28] & 385 & 3/6 \\
 & Dialogue line length~$\sim$ & -1.11$^{*}$ & [-1.50, -0.74] & 382 & 5/6 \\
 & Voice-over cues~$\sim$ & -0.01 & [-0.10, +0.09] & 385 & 4/6 \\
 & Unfilmable action lines~$\downarrow$ & -0.30$^{*}$ & [-0.40, -0.20] & 385 & 6/6 \\
 & Stimulus words~$\uparrow$ & -1.16$^{*}$ & [-1.92, -0.42] & 385 & 5/6 \\
 & Plot conflicts / 1k words~$\uparrow$ & +1.36$^{*}$ & [+1.13, +1.59] & 385 & 6/6 \\
 & Hard-cut ending rate~$\uparrow$ & +13.3\,pt$^{*}$ & [+8.1, +18.5] & 384 & 6/6 \\
 & Cold-open rate~$\uparrow$ & +5.2\,pt$^{*}$ & [+1.3, +9.1] & 384 & 5/6 \\
 & Prop-activation rate~$\uparrow$ & +0.0\,pt & [-6.2, +6.2] & 209 & 1/6 \\
 & Ending-twist rate~$\sim$ & +2.9\,pt & [-1.6, +7.5] & 385 & 4/6 \\
\midrule
K2.6: Sonnet library vs Base & Longest dialogue volley~$\sim$ & +0.51$^{*}$ & [+0.34, +0.67] & 385 & 5/6 \\
 & Dialogue line length~$\sim$ & -1.01$^{*}$ & [-1.33, -0.68] & 385 & 6/6 \\
 & Voice-over cues~$\sim$ & +0.24$^{*}$ & [+0.11, +0.37] & 385 & 5/6 \\
 & Unfilmable action lines~$\downarrow$ & -0.22$^{*}$ & [-0.32, -0.12] & 385 & 6/6 \\
 & Stimulus words~$\uparrow$ & +1.57$^{*}$ & [+0.57, +2.54] & 385 & 6/6 \\
 & Plot conflicts / 1k words~$\uparrow$ & +1.02$^{*}$ & [+0.80, +1.25] & 385 & 6/6 \\
 & Hard-cut ending rate~$\uparrow$ & +5.2\,pt$^{*}$ & [+0.5, +10.2] & 384 & 6/6 \\
 & Cold-open rate~$\uparrow$ & +6.2\,pt$^{*}$ & [+2.3, +10.2] & 384 & 5/6 \\
 & Prop-activation rate~$\uparrow$ & +6.9\,pt & [-0.5, +13.9] & 202 & 5/6 \\
 & Ending-twist rate~$\sim$ & +0.0\,pt & [-4.2, +4.2] & 385 & 1/6 \\
\midrule
Opus 4.8: own library vs Base & Longest dialogue volley~$\sim$ & -0.23$^{*}$ & [-0.34, -0.11] & 385 & 5/6 \\
 & Dialogue line length~$\sim$ & +0.28 & [-0.08, +0.64] & 385 & 3/6 \\
 & Voice-over cues~$\sim$ & +0.45$^{*}$ & [+0.31, +0.60] & 385 & 5/6 \\
 & Unfilmable action lines~$\downarrow$ & -0.29$^{*}$ & [-0.38, -0.20] & 385 & 6/6 \\
 & Stimulus words~$\uparrow$ & -0.48$^{*}$ & [-0.88, -0.09] & 385 & 5/6 \\
 & Plot conflicts / 1k words~$\uparrow$ & +0.58$^{*}$ & [+0.43, +0.74] & 385 & 6/6 \\
 & Hard-cut ending rate~$\uparrow$ & +27.0\,pt$^{*}$ & [+21.8, +32.2] & 385 & 6/6 \\
 & Cold-open rate~$\uparrow$ & +6.0\,pt$^{*}$ & [+2.9, +9.1] & 385 & 5/6 \\
 & Prop-activation rate~$\uparrow$ & +3.1\,pt & [-2.7, +8.9] & 225 & 2/6 \\
 & Ending-twist rate~$\sim$ & +3.4\,pt & [-0.8, +7.5] & 385 & 5/6 \\
\midrule
Opus 4.8: Sonnet library vs Base & Longest dialogue volley~$\sim$ & -0.23$^{*}$ & [-0.34, -0.11] & 385 & 6/6 \\
 & Dialogue line length~$\sim$ & +0.03 & [-0.28, +0.35] & 385 & 3/6 \\
 & Voice-over cues~$\sim$ & +0.05 & [-0.09, +0.18] & 385 & 4/6 \\
 & Unfilmable action lines~$\downarrow$ & -0.40$^{*}$ & [-0.49, -0.31] & 385 & 6/6 \\
 & Stimulus words~$\uparrow$ & +1.11$^{*}$ & [+0.65, +1.56] & 385 & 6/6 \\
 & Plot conflicts / 1k words~$\uparrow$ & +1.05$^{*}$ & [+0.89, +1.21] & 385 & 6/6 \\
 & Hard-cut ending rate~$\uparrow$ & +22.1\,pt$^{*}$ & [+17.1, +27.0] & 385 & 6/6 \\
 & Cold-open rate~$\uparrow$ & +10.9\,pt$^{*}$ & [+7.3, +14.5] & 385 & 6/6 \\
 & Prop-activation rate~$\uparrow$ & +14.7\,pt$^{*}$ & [+8.4, +20.9] & 225 & 6/6 \\
 & Ending-twist rate~$\sim$ & +2.1\,pt & [-2.3, +6.5] & 385 & 4/6 \\
\midrule
Sonnet 4.6: +Distill vs Base & Longest dialogue volley~$\sim$ & +0.42$^{*}$ & [+0.21, +0.64] & 385 & 6/6 \\
 & Dialogue line length~$\sim$ & +1.50$^{*}$ & [+1.13, +1.87] & 384 & 4/6 \\
 & Voice-over cues~$\sim$ & +2.21$^{*}$ & [+1.91, +2.51] & 385 & 6/6 \\
 & Unfilmable action lines~$\downarrow$ & +0.30$^{*}$ & [+0.14, +0.46] & 385 & 6/6 \\
 & Stimulus words~$\uparrow$ & +0.04 & [-0.25, +0.33] & 385 & 4/6 \\
 & Plot conflicts / 1k words~$\uparrow$ & -0.28$^{*}$ & [-0.40, -0.17] & 385 & 5/6 \\
 & Hard-cut ending rate~$\uparrow$ & -2.1\,pt & [-7.3, +2.9] & 385 & 3/6 \\
 & Cold-open rate~$\uparrow$ & +0.8\,pt & [-3.4, +4.9] & 385 & 3/6 \\
 & Prop-activation rate~$\uparrow$ & +2.8\,pt & [-3.3, +8.8] & 215 & 4/6 \\
 & Ending-twist rate~$\sim$ & -1.3\,pt & [-6.0, +3.1] & 385 & 4/6 \\
\midrule
K2.6: own library vs Sonnet library & Longest dialogue volley~$\sim$ & -0.39$^{*}$ & [-0.58, -0.20] & 385 & 5/6 \\
 & Dialogue line length~$\sim$ & -0.09 & [-0.38, +0.20] & 382 & 4/6 \\
 & Voice-over cues~$\sim$ & -0.24$^{*}$ & [-0.38, -0.11] & 385 & 6/6 \\
 & Unfilmable action lines~$\downarrow$ & -0.08 & [-0.16, +0.01] & 385 & 3/6 \\
 & Stimulus words~$\uparrow$ & -2.74$^{*}$ & [-3.58, -1.92] & 385 & 6/6 \\
 & Plot conflicts / 1k words~$\uparrow$ & +0.34$^{*}$ & [+0.07, +0.60] & 385 & 5/6 \\
 & Hard-cut ending rate~$\uparrow$ & +8.1\,pt$^{*}$ & [+2.3, +13.5] & 385 & 6/6 \\
 & Cold-open rate~$\uparrow$ & -1.0\,pt & [-4.7, +2.6] & 385 & 2/6 \\
 & Prop-activation rate~$\uparrow$ & -2.6\,pt & [-8.8, +3.5] & 227 & 3/6 \\
 & Ending-twist rate~$\sim$ & +2.9\,pt & [-1.8, +7.5] & 385 & 5/6 \\
\midrule
Opus 4.8: own library vs Sonnet library & Longest dialogue volley~$\sim$ & +0.00 & [-0.10, +0.10] & 385 & 0/6 \\
 & Dialogue line length~$\sim$ & +0.25 & [-0.06, +0.57] & 385 & 2/6 \\
 & Voice-over cues~$\sim$ & +0.40$^{*}$ & [+0.27, +0.54] & 385 & 5/6 \\
 & Unfilmable action lines~$\downarrow$ & +0.11$^{*}$ & [+0.04, +0.18] & 385 & 6/6 \\
 & Stimulus words~$\uparrow$ & -1.59$^{*}$ & [-2.03, -1.16] & 385 & 6/6 \\
 & Plot conflicts / 1k words~$\uparrow$ & -0.47$^{*}$ & [-0.62, -0.31] & 385 & 6/6 \\
 & Hard-cut ending rate~$\uparrow$ & +4.9\,pt & [-0.3, +9.9] & 385 & 4/6 \\
 & Cold-open rate~$\uparrow$ & -4.9\,pt$^{*}$ & [-8.1, -2.1] & 385 & 5/6 \\
 & Prop-activation rate~$\uparrow$ & -13.9\,pt$^{*}$ & [-19.9, -8.2] & 231 & 6/6 \\
 & Ending-twist rate~$\sim$ & +1.3\,pt & [-2.6, +5.2] & 385 & 3/6 \\
\end{longtable}
\endgroup

}

\clearpage

\section{Held-Out Metrics and Library Composition Across Models}\label{sec:appendix:xmodel}

This appendix supports the cross-model experiments in \Cref{sec:exp:xmodel}.
We first report how each model performs on the held-out novels with each library, and then compare what the Sonnet and Opus libraries contain to explain why Opus benefits more from Sonnet's library than from its own.

\subsection{Held-Out Metrics}\label{sec:appendix:xmodel_metrics}

\Cref{tab:xmodel_full} provides the ten-metric results for the experiments in \Cref{sec:exp:xmodel} on the same six held-out novels.
Each model is evaluated with an empty library, its own five-novel library, and Sonnet's five-novel library transferred without modification.

\begin{table}[h]
\centering
\caption{Full cross-model results with empty, own-model, and transferred Sonnet libraries trained on five novels. Value columns report mean counts or densities and binary yes-rates. Bold marks the highest $\uparrow$ or lowest $\downarrow$ value among configurations; $\sim$ denotes descriptive metrics. $^{*}$ marks raw paired differences from Base whose 95\% bootstrap intervals exclude zero. Auxiliary $d$ columns report paired MAE against human originals or model-minus-human rate gaps in percentage points.}
\label{tab:xmodel_full}
\footnotesize
\setlength{\tabcolsep}{2.5pt}
\begin{tabular*}{\linewidth}{@{\extracolsep{\fill}}lrrrrrrr@{}}
\toprule
Metric & Base & +Own library & +Sonnet library & Human & $d_{\text{Base}}$ & $d_{\text{Own}}$ & $d_{\text{Sonnet}}$ \\
\midrule
\multicolumn{8}{@{}l}{\textbf{Kimi K2.6 as base model}} \\
\rowcolor{tabrule}\multicolumn{8}{@{}l}{\textit{Line-level style}} \\
Longest dialogue volley (turns/ep)~$\sim$ & 3.0 & 3.1 & 3.5$^{*}$ & 3.6 & 1.45 & 1.66 & 1.59 \\
Dialogue line length (words)~$\sim$ & 7.6 & 6.5$^{*}$ & 6.6$^{*}$ & 11.9 & 4.80 & 5.56 & 5.42 \\
Voice-over cues (per ep)~$\sim$ & 0.37 & 0.36 & 0.60$^{*}$ & 0.22 & 0.43 & 0.44 & 0.63 \\
Unfilmable action lines (per ep)~$\downarrow$ & 0.5 & \textbf{0.2}$^{*}$ & 0.3$^{*}$ & 0.4 & 0.63 & 0.45 & 0.49 \\
Stimulus words (per ep)~$\uparrow$ & 9.6 & 8.4$^{*}$ & \textbf{11.2}$^{*}$ & 9.2 & 5.05 & 4.79 & 5.75 \\
\midrule
\rowcolor{tabjudge}\multicolumn{8}{@{}l}{\textit{Episode-level structure}} \\
Plot conflicts (per 1k words)~$\uparrow$ & 2.89 & \textbf{4.25}$^{*}$ & 3.91$^{*}$ & 3.47 & 1.95 & 2.27 & 2.23 \\
Hard-cut ending rate~$\uparrow$ & 32\% & \textbf{45\%}$^{*}$ & 37\%$^{*}$ & 62\% & $-30.6$ & $-17.1$ & $-25.2$ \\
Cold-open rate~$\uparrow$ & 69\% & 74\%$^{*}$ & \textbf{75\%}$^{*}$ & 59\% & $+10.0$ & $+15.3$ & $+16.4$ \\
Prop-activation rate~$\uparrow$ & 56\% & 58\% & \textbf{60\%} & 57\% & $-1.7$ & $+0.9$ & $+3.0$ \\
Ending-twist rate~$\sim$ & 31\% & 34\% & 31\% & 31\% & $-0.3$ & $+2.6$ & $-0.3$ \\
\midrule
\multicolumn{8}{@{}l}{\textbf{Opus 4.8 as base model}} \\
\rowcolor{tabrule}\multicolumn{8}{@{}l}{\textit{Line-level style}} \\
Longest dialogue volley (turns/ep)~$\sim$ & 3.3 & 3.0$^{*}$ & 3.0$^{*}$ & 3.6 & 1.38 & 1.36 & 1.38 \\
Dialogue line length (words)~$\sim$ & 6.4 & 6.7 & 6.5 & 11.9 & 5.72 & 5.63 & 5.76 \\
Voice-over cues (per ep)~$\sim$ & 0.66 & 1.12$^{*}$ & 0.71 & 0.22 & 0.69 & 1.04 & 0.66 \\
Unfilmable action lines (per ep)~$\downarrow$ & 0.6 & 0.3$^{*}$ & \textbf{0.2}$^{*}$ & 0.4 & 0.62 & 0.53 & 0.44 \\
Stimulus words (per ep)~$\uparrow$ & 8.8 & 8.3$^{*}$ & \textbf{9.9}$^{*}$ & 9.2 & 4.19 & 4.12 & 4.52 \\
\midrule
\rowcolor{tabjudge}\multicolumn{8}{@{}l}{\textit{Episode-level structure}} \\
Plot conflicts (per 1k words)~$\uparrow$ & 2.59 & 3.17$^{*}$ & \textbf{3.64}$^{*}$ & 3.47 & 1.97 & 1.94 & 1.85 \\
Hard-cut ending rate~$\uparrow$ & 46\% & \textbf{73\%}$^{*}$ & 68\%$^{*}$ & 62\% & $-16.1$ & $+10.9$ & $+6.0$ \\
Cold-open rate~$\uparrow$ & 68\% & 74\%$^{*}$ & \textbf{79\%}$^{*}$ & 59\% & $+9.4$ & $+15.3$ & $+20.3$ \\
Prop-activation rate~$\uparrow$ & 56\% & 60\% & \textbf{73\%}$^{*}$ & 57\% & $-1.2$ & $+2.3$ & $+15.5$ \\
Ending-twist rate~$\sim$ & 40\% & 43\% & 42\% & 31\% & $+8.6$ & $+11.9$ & $+10.6$ \\
\bottomrule
\end{tabular*}

\end{table}

\subsection{Library Composition}\label{sec:appendix:xmodel_library}

The performance divergence between Opus's own library and the transferred Sonnet library discussed in \Cref{sec:exp:xmodel} aligns with differences in their structural composition. Since the training loop is largely error-driven, Opus's strong base capabilities may lead to a library that focuses more on niche, genre-specific constraints rather than foundational rules.

\Cref{tab:library_composition} shows that Opus dedicates 63\% of its library text to genre-specific cards against 43\% for Sonnet. Opus's library contains fewer general quantitative rules, defined as sentences containing both a number with a unit and a constraint word like \enquote{no more than}. It generates 0.13 quantitative rules per thousand characters, whereas Sonnet generates 0.67 rules per thousand characters as it continually corrects basic errors.

During training, Sonnet continuously refines these foundational rules and revises each card a median of 19 times against 4 times for Opus. This statistical breakdown suggests that although Opus is a superior generator, its lower frequency of foundational errors might make it a less effective rule-miner for core writing techniques.

\begin{table}[h]
\centering
\caption{Composition of the libraries trained by Sonnet 4.6 and Opus 4.8 on five novels. General cards include both the general-principle and the craft-technique folders.}
\label{tab:library_composition}
\small
\begin{tabular}{@{}lrr@{}}
\toprule
 & Sonnet 4.6 & Opus 4.8 \\
\midrule
Steps / micro-steps / library commits & 42 / 75 / 76 & 42 / 75 / 76 \\
Cards (general / genre-specific) & 17 (10 / 7) & 27 (11 / 16) \\
Genre-specific share of text & 43\% & 63\% \\
Quantitative rules per 1k characters & 0.67 & 0.13 \\
Median revisions per card & 19 & 4 \\
Genre-specific cards read per step (Opus) & 2.5 & 4.2 \\
General cards read per step (Opus) & 9.1 & 8.8 \\
\bottomrule
\end{tabular}
\end{table}
\clearpage

\section{Case Study: Transfer into an External Production Pipeline}\label{sec:appendix:case}

As introduced in \Cref{sec:exp:case}, we deploy the learned skill library in a third-party production pipeline that outputs tabular storyboards rather than Hollywood-format screenplays. This appendix provides the detailed qualitative evidence behind that experiment. We generated five episodes of an unseen title, \emph{Second Bloom} (anonymized), under identical conditions: same story outline, prompt template, and base model (Claude Sonnet 4.6). The treatment group had access to the skill library and could retrieve individual rule cards on demand, while the control group operated with an empty skill library. A per-episode read trace records which cards the treatment agent actually read.

The generated outputs are tabular shot lists with scene numbers, visual descriptions, and dialogue columns, demonstrating that the agent applies narrative skills to a camera-ready format it never encountered during training. Dialogue and reasoning traces below are quoted verbatim with names substituted. The tag [spectacle] represents the pipeline's marker for a visually striking beat.

\paragraph{Enforcing Ending Discipline (Episode 1).}
The outline establishes a clear cliffhanger: the husband sneers that his wife will return crawling within two days, oblivious to the fact that she is removing her wedding ring. \Cref{tab:app-ep01-beats} aligns the shot budgets. The two groups track each other beat by beat leading up to the climax, but the control group tacks on five extra shots after the ring lands. \Cref{tab:app-ep01-endings} prints both endings. The control agent correctly identified the hook in its reasoning trace:

\begin{quote}\small\itshape
The key hook for the next episode is that Marcus doesn't see Nora remove the ring while he's laughing coldly about her coming back in a couple days---that moment of him being oblivious while she's already made her move is what makes the ending work.
\end{quote}

Yet the shipped control storyboard resolves the suspense it just named: shots 26--30 linger on the ring, give Marcus a spoken reaction and a closing voice-over, and let Nora exit the room. The treatment agent read the ending rule card (one of seven cards in its read trace) and locked the cut on the climactic action:

\begin{quote}\small\itshape
For the episode ending, I need to follow the hook structure---cutting right on the climactic action without resolution. [\dots] the sequence flows: Marcus delivers the cold dismissal in shot 23, Nora responds and removes the ring in shot 24, then the ring lands on the table in shot 25 as the final cut. That's the right structure. The ring on the table becomes the hook for the next episode---Marcus doesn't notice it yet, but the audience does, creating that information gap that makes them wonder what happens when he realizes.
\end{quote}

The episode stops on shot 25, on the clink of the ring. Here the rule card contributed discipline, not ideas: both agents recognized the hook, but only the agent holding the rule preserved it.

\begin{table}[h]
\centering
\caption{Episode 1, shot budget by story beat. The two groups match almost beat for beat before the climax. The length gap is the post-climax tail.}
\label{tab:app-ep01-beats}
\footnotesize
\begin{tabular}{lcc}
\toprule
Story beat & Control (30 shots) & Treatment (25 shots) \\
\midrule
Corridor cold open & 1--5 (5) & 1--5 (5) \\
Flashback: the divorce demand & 6--9 (4) & 6--10 (5) \\
Return to the present & 10 (1) & 11 (1) \\
Mediation room, before the declaration & 11--15 (5) & 12--16 (5) \\
Calm declaration; the room freezes & 16--18 (3) & 17--19 (3) \\
Confrontation & 19--22 (4) & 20--23 (4) \\
Ring removal & 23--25 (3) & 24--25 (2) \\
Tail after the climax & 26--30 (5) & none \\
\bottomrule
\end{tabular}
\end{table}

\begin{table}[h]
\centering
\caption{Episode 1, the two endings. Dialogue verbatim except names. The control group plays five more shots after the ring lands. The treatment group cuts on the clink.}
\label{tab:app-ep01-endings}
\footnotesize
\begin{minipage}[t]{0.48\textwidth}
\raggedright
\textbf{Control, shots 25--30 (episode ends on 30)}\par\medskip
\textbf{S25}~(close-up)\enspace Nora lays the wedding ring gently on the table edge. Her fingers let go and withdraw.\par
\emph{SFX: Soft clink on wood.}\enspace {[}spectacle: ring removed in front of everyone; all eyes converge{]}\par\smallskip
\textbf{S26}~(wide)\enspace Everyone in the mediation room turns to the ring on the table edge. No one speaks.\par
\emph{SFX: Dead silence.}\par\smallskip
\textbf{S27}~(close)\enspace Marcus stares at the ring on the table edge, his face blank.\par
\emph{Marcus: (low) You're not serious right now.}\par\smallskip
\textbf{S28}~(close)\enspace Nora picks up her bag and stands.\par
\emph{Nora: Have your lawyer contact mine, Marcus.}\par\smallskip
\textbf{S29}~(medium)\enspace Nora walks to the door, her back to the room, her steps steady.\par
\emph{SFX: Footsteps receding.}\par\smallskip
\textbf{S30}~(close-up)\enspace Marcus reaches out, picks the ring up from the table edge, and closes his hand around it.\par
\emph{Marcus: (V.O.) She'll be back. She always comes back.}
\end{minipage}\hfill
\begin{minipage}[t]{0.48\textwidth}
\raggedright
\textbf{Treatment, shots 24--25 (episode ends on 25)}\par\medskip
\textbf{S24}~(close-up)\enspace Nora's fingers touch the wedding ring, slide it slowly off her ring finger, and hold it in her palm.\par
\emph{Nora: Then I'll have nothing.}\enspace {[}spectacle{]}\par\smallskip
\textbf{S25}~(close-up)\enspace The wedding ring is laid gently on the table edge with a crisp metal clink. Nora's hand lets go and withdraws. Marcus still has his back to the table.\par
\emph{SFX: Wedding ring placed on table with a soft clink.}\enspace {[}spectacle: episode-ending hook{]}\par\medskip
\emph{(The episode ends here.)}
\end{minipage}
\end{table}

\paragraph{Translating Voice-over into Action (Episode 2).}
This episode presents the heaviest narration trap of the five: the protagonist returns to an empty mansion and watches seven years of her marriage decay in flashback. With no scene partner for most of the episode, the easiest way to fill the line quota is voice-over. \Cref{tab:app-ep02-counts} details the line distribution: the control agent narrates 15 of its 24 lines, burying 6 of them inside flashbacks. The treatment agent retains only four voice-over lines, none within flashbacks, and actually generates more spoken dialogue than the control.

\Cref{tab:app-ep02-beat} sets a critical beat side by side: the daughter runs past her mother and into the arms of the mistress. The control agent handles the pain through narration. The mother grips the tray while three voice-over lines restate what the frame already shows. The treatment agent, guided by a voice-over rule (act out emotions instead of narrating them) and a prop rule (a prop must be set off, not clutched), dramatizes the rejection entirely. The mother is cut off mid-sentence, the child verbally rejects her, and the tray tips until the cookies shatter on the floor.

\begin{table}[h]
\centering
\caption{Episode 2, line counts. With the same outline, the two groups spend their lines differently.}
\label{tab:app-ep02-counts}
\small
\begin{tabular}{lcc}
\toprule
 & Control & Treatment \\
\midrule
Shots & 30 & 25 \\
Dialogue + voice-over lines & 24 & 18 \\
Voice-over lines & 15 (62.5\%) & 4 (22\%) \\
Voice-over inside flashbacks & 6 & 0 \\
Spoken lines & 9 & 14 \\
\bottomrule
\end{tabular}
\end{table}

\begin{table}[h]
\centering
\caption{Episode 2, the same rejection beat in both groups. The control group narrates the event while the treatment group plays it out through action.}
\label{tab:app-ep02-beat}
\footnotesize
\begin{minipage}[t]{0.48\textwidth}
\raggedright
\textbf{Control, shots 19--22}\par
\emph{(In the preceding shot, Serena kneels and hands Mia a gift.)}\par\medskip
\textbf{S19}~(wide)\enspace Mia throws herself at Serena, arms around her neck. Nora stands at the kitchen door holding a tray.\par
\emph{Mia: I love it! Thank you, Auntie Serena!}\enspace {[}spectacle{]}\par\smallskip
\textbf{S20}~(close)\enspace Nora grips the tray, her eyes on Mia and Serena.\par
\emph{Nora: (V.O.) My daughter hadn't hugged me like that in months.}\par\smallskip
\textbf{S21}~(medium)\enspace Mia takes Serena's hand and runs toward the bedroom, passing Nora without a glance.\par
\emph{Nora: (V.O.) She walked right past me like I was furniture.}\par\smallskip
\textbf{S22}~(close-up)\enspace Nora sets the tray down and smooths her apron.\par
\emph{Nora: (V.O.) That's when I knew. I wasn't her mother anymore. I was just the help.}
\end{minipage}\hfill
\begin{minipage}[t]{0.48\textwidth}
\raggedright
\textbf{Treatment, shots 11--14}\par\medskip
\textbf{S11}~(medium)\enspace Mia runs down the stairs, straight past Nora, and throws herself into Serena's arms; Nora stands in place holding a tray.\par
\emph{Mia: Serena! I missed you so much!}\par\smallskip
\textbf{S12}~(close)\enspace Nora steps forward and holds the tray out to Mia.\par
\emph{Nora: Mia --- I made your favorite cookies---}\par\smallskip
\textbf{S13}~(close)\enspace Mia turns her back on Nora and takes Serena's hand.\par
\emph{Mia: I want Serena to tuck me in tonight. Not you.}\par\smallskip
\textbf{S14}~(close-up)\enspace Nora's hand holds the tray. It tips, and the cookies slide off the edge and shatter on the floor.\par
\emph{SFX: Plate tilting and cookies crashing to the floor.}\enspace {[}spectacle{]}
\end{minipage}
\end{table}

\paragraph{In-Context Self-Correction (Episode 3).}
This case closes the mechanism loop end to end: a rule is read, a violation is caught, the draft is revised, and the shipped artifact changes, with every step of this process fully recorded in the logs. The treatment agent loaded the ending rule card before drafting (one of four cards in its read trace). During self-review, it interrupted its own process:

\begin{quote}\small\itshape
I'm checking the shot count and realizing my ending violates the soft landing rule I noted earlier. The memory specifically warns against ending on V.O., gazing, or static moments---I need a decisive physical action instead. So I should reframe shot 27 to show Nora actually clicking the button or picking up her phone to call, something that propels the story forward rather than leaving her frozen in contemplation.
\end{quote}

The shipped storyboard ends exactly as revised. Its final shot:

\begin{quote}\small
\textbf{S27}~(medium; Vance residence, master bedroom, late night)\enspace Nora turns back to the screen. Over her shoulder, the laptop page reads \enquote{The Old Harlan Place --- Greenhouse For Sale}. Her hand reaches for the trackpad and presses. \emph{SFX: Mouse click.}\enspace Remark: episode-ending hook; Nora clicks to book a visit.
\end{quote}

\paragraph{Limitation: Skill Activation Failure (Episode 5).}
Possessing a useful rule does not guarantee it will be applied. Over the five episodes, spontaneous card reading steadily declined from seven cards in Episode 1 to three in Episode 5. In this episode, the agent did not activate the ending rule, the only episode in which that card was missed.
The resulting script contains 17 voice-over lines out of 21 dialogue lines, compared with ten in the control, and ends on a static fade-out rather than a decisive action. The final shot shows morning light falling on the note she left on the nightstand.
Together with the successful self-correction in Episode 3, this failure highlights that reliable skill activation is a necessary condition for the learned rules to take effect.

\section{Held-Out Metrics for Corpus Scaling}\label{sec:appendix:scaling}

\Cref{tab:scaling_full} provides the complete held-out results for the corpus-size analysis in \Cref{sec:exp:scaling}.
All libraries are trained with Sonnet 4.6 and evaluated on the same six held-out novels.
Bold marks the highest $\uparrow$ or lowest $\downarrow$ value among configurations; $\sim$ metrics are descriptive and have no bold winner.

\begin{table}[h]
\centering
\caption{Corpus-size comparison on held-out novels. Count and density rows report group means with same-episode human MAE as $d$. Rate rows report yes-rates with signed percentage-point gaps to human as $d$. In the value columns, $^{*}$ marks paired differences from Base whose 95\% bootstrap intervals exclude zero, using raw counts, densities, or binary verdicts. The $d$ columns are supplementary comparisons with human originals.}
\label{tab:scaling_full}
\small
\setlength{\tabcolsep}{2.5pt}
\resizebox{\linewidth}{!}{%
\begin{tabular}{lrrrr>{\columncolor{black!6}}rrrrr}
\toprule
Metric & Base & +Ours 5n & +Ours 10n & +Ours 20n & Human & $d_{\text{Base}}$ & $d_{\text{5n}}$ & $d_{\text{10n}}$ & $d_{\text{20n}}$ \\
\midrule
\rowcolor{tabrule}\multicolumn{10}{l}{\textit{Line-level style}} \\
Longest dialogue volley (turns/ep)~$\sim$ & 4.46 & 4.03$^{*}$ & 4.28 & 4.18$^{*}$ & 3.60 & 1.86 & 1.67 & 1.81 & 1.72 \\
Dialogue line length (words)~$\sim$ & 5.63 & 6.99$^{*}$ & 8.21$^{*}$ & 7.34$^{*}$ & 11.89 & 6.53 & 5.38 & 4.49 & 4.99 \\
Voice-over cues (per ep)~$\sim$ & 0.35 & 0.40 & 1.09$^{*}$ & 0.52$^{*}$ & 0.22 & 0.48 & 0.48 & 1.02 & 0.58 \\
Unfilmable action lines (per ep)~$\downarrow$ & 1.19 & 0.46$^{*}$ & 0.54$^{*}$ & \textbf{0.17}$^{*}$ & 0.36 & 1.09 & 0.60 & 0.63 & 0.46 \\
Stimulus words (per ep)~$\uparrow$ & 3.26 & 5.93$^{*}$ & \textbf{6.62}$^{*}$ & 6.28$^{*}$ & 9.24 & 6.54 & 4.85 & 4.84 & 5.01 \\
\midrule
\rowcolor{tabjudge}\multicolumn{10}{l}{\textit{Episode-level structure}} \\
Plot conflicts (per 1k words)~$\uparrow$ & 2.10 & 3.09$^{*}$ & 2.93$^{*}$ & \textbf{3.69}$^{*}$ & 3.47 & 2.17 & 1.99 & 2.03 & 2.06 \\
Hard-cut ending rate~$\uparrow$ & 34\% & 56\%$^{*}$ & 45\%$^{*}$ & \textbf{67\%}$^{*}$ & 62\% & $-28.3$ & $-6.0$ & $-17.4$ & $+4.7$ \\
Cold-open rate~$\uparrow$ & 52\% & 69\%$^{*}$ & 72\%$^{*}$ & \textbf{74\%}$^{*}$ & 59\% & $-7.3$ & $+10.1$ & $+13.2$ & $+15.1$ \\
Prop-activation rate~$\uparrow$ & 46\% & 58\%$^{*}$ & \textbf{59\%}$^{*}$ & 58\%$^{*}$ & 57\% & $-11.4$ & $+0.5$ & $+1.2$ & $+0.8$ \\
Ending-twist rate~$\sim$ & 33\% & 38\%$^{*}$ & 39\%$^{*}$ & 31\% & 31\% & $+1.6$ & $+6.5$ & $+7.5$ & $-0.8$ \\
\bottomrule
\end{tabular}}

\end{table}
\clearpage

\section{Skill Library Content Across Corpus Sizes}\label{sec:appendix:library_content}

While \Cref{sec:exp:scaling} evaluates the libraries quantitatively through held-out metrics, this section investigates \textit{what} specific skills the agent actually acquires as the training corpus grows. To analyze this evolution, we decomposed each library's cards into atomic writing instructions and cross-referenced them to identify semantic overlaps. Claude Opus 4.8 decomposed the five-, ten-, and twenty-novel libraries into 337, 290, and 472 atomic rules and matched every rule against the other libraries. To check this matching, GPT-5.5 independently re-matched the 472 twenty-novel rules against the five-novel library, and the two judges agree on whether a rule has any counterpart for 83\% of them. \Cref{tab:library_content} highlights how the acquired rules progress across the three corpus sizes.

\begin{table}[h]
\centering
\caption{Representative rules in the five-novel library and rules added at ten and twenty novels. An added rule has no full or partial counterpart in any smaller library. The rules presented here are anonymized versions that replace examples tied to specific source stories with generic ones and are shortened for brevity.}
\label{tab:library_content}
\small
\renewcommand{\arraystretch}{1.15}
\begin{tabular}{@{}p{0.32\linewidth}p{0.32\linewidth}p{0.32\linewidth}@{}}
\toprule
Five novels & Added at ten novels & Added at twenty novels \\
\midrule
Delete idle beats such as staring, walking, or musing. &
When a character says the opposite of what they feel, reveal the true feeling in a voice-over on the same beat. &
Stage an identity reveal in four beats: a bystander doubts, an authority suppresses, the lead reveals the truth in one line, and the room collapses. \\
\addlinespace
Turn every emotion into a visible action, such as smashing a cup, rather than reddening eyes. &
Use at most two voice-overs per episode. &
End a face-slap with a boomerang line that throws the villain's insult back. \\
\addlinespace
Allow no more than three turns of pure dialogue in a standoff before a physical escalation. &
A child character exists to push the leads into close contact or to say what the adults will not. &
A scheming rival wins through guilt and family pressure rather than direct orders. \\
\addlinespace
State the intent in one line and start the next line at the destination. &
Hold one emotion for no more than three lines before switching to its opposite. &
An extreme action needs an emotional cause, and the leads unite as soon as the truth comes out. \\
\addlinespace
End on the bombshell line and cut to black without showing the reaction. &
Mark emotion in parentheses only before an emotional peak, four to six times per episode. &
Use no more than three scene headings per episode. \\
\bottomrule
\end{tabular}
\end{table}

\paragraph{Five novels: Core Craft and Line-Level Execution.}
With just five novels, the library successfully captures the foundational mechanics of the medium. The rules primarily focus on the direct execution of scenes, instructing the agent to eliminate internal monologues, remove idle beats, and translate abstract emotions into visible actions (e.g., smashing a cup instead of reddening eyes). It also establishes basic pacing principles, such as breaking up long dialogue exchanges with physical actions and cutting to black immediately after a bombshell line.

\paragraph{Ten novels: Voice-Overs and Supporting Roles.}
Expanding the corpus to ten novels gives voice-overs and supporting characters specific dramatic functions. The ten-novel library adds a dedicated voice-over card. It lists situations where a voice-over should reveal what a character hides, such as a line that says the opposite of what the character feels, and it caps voice-overs at two per episode. Because the average use stays well below this cap, the new card mainly encourages voice-overs, in line with the rise in voice-over use reported in \Cref{sec:exp:scaling}. Furthermore, the agent begins to assign structural purposes to supporting characters. For example, a child character exists specifically to force the protagonists into close proximity. This demonstrates an emerging grasp of relationship dynamics beyond the two leads.

\paragraph{Twenty novels: Macro-Structure and Complex Tactics.}
As the corpus scales to twenty novels, the acquired skills shift decisively toward episode-level architecture and complex character strategies. Standard narrative tropes, such as \enquote{face-slapping} or identity reveals, evolve into highly structured, step-by-step templates. Characters also develop more sophisticated interaction tactics, learning to manipulate rivals through guilt and family pressure rather than relying on direct commands. Additionally, the library enforces stricter scene economy, explicitly limiting the number of locations per episode to maintain narrative momentum.

\paragraph{Shared Foundations and Refinement.}
Although each library is trained from an empty library, larger corpora keep the foundations that smaller ones learn. 295 of the 337 rules in the five-novel library, or 88\%, have a full or partial counterpart in the twenty-novel library. Rather than simply stacking disjointed rules, the larger library refines them: broad instructions in the smaller library tend to become more precise, context-aware procedures. This qualitative shift, from mastering line-level mechanics to orchestrating entire episodes, mirrors the quantitative training dynamics observed in \Cref{sec:exp:dynamics}, where line-level style converges quickly while episode-level structure requires sustained training to improve.

\raggedbottom  %
\paragraph{Example Rule Cards.}
To show what the script agent actually reads, we reproduce three rule cards from the twenty-novel library used by +Ours, one from each expert folder. Each card opens with a description, which the agent reads when deciding whether to open the card, followed by the rules themselves. We show the description and the first few rules of each card and omit the rest for space. The token counts refer to the full cards, all of which stay within the limits described in \Cref{sec:method:memory}. The reproduced cards are anonymized and shortened versions where specific story references have been replaced with generic examples.

\begin{rulecard}{General principles \quad \texttt{short-drama-medium-fundamentals.md}}{description 106 tokens, rules 510 tokens}

\textit{Description.} Read before writing any script. This is the foundation for every writing decision. It defines what vertical short drama is as a medium, how its audience watches it, and what a good script means in this medium. It does not apply to feature films or long-form series.

Vertical short drama is an emotion-delivery machine, not a work of storytelling art. The viewer's thumb hovers over the screen, ready to swipe away, and retention is counted in seconds rather than episodes.

\textbf{The medium.} Every five seconds must give the viewer a reason to keep watching, such as a scream, a slap, a thrown drink, or an exposed secret. A boring frame means the viewer is gone.

\textbf{Extreme time compression.} Enter every scene at its peak tension and delete all entrance setup, such as hallways, waiting, opening doors, sitting down, or small talk in a car. The same holds between scenes. One emotional peak cuts straight into the next event, with no transition.

\textbf{How viewers watch.} Zero cognitive barrier. Viewers do not think, they feel. Complex subtext, subtle character arcs, and careful foreshadowing do not work here. Only emotional stimuli that land instantly do.

\textit{[Remaining rules omitted.]}
\end{rulecard}

\begin{rulecard}{Craft techniques \quad \texttt{prop-function.md}}{description 192 tokens, rules 1{,}144 tokens}

\textit{Description.} Read whenever you choose a concrete prop or object for a scene, whether it serves conflict, humiliation, a display of wealth, a declaration of love, a reveal, or exposed evidence. You must read it before using a USB drive, a screen cast, a screenshot, a document, a legal file, or high-tech gear to reveal evidence or move the plot, because such props are almost always wrong. The model tends to pick stereotyped upscale props, while human writers pick cheap everyday objects to create spectacle. It does not apply to purely verbal arguments. For physical contact, see \texttt{conflict-physical-action.md}.

\textbf{Emotional breakdowns use cheap props in action.} An angry character grabs something cheap and acts at once, gulping a drink or smashing a phone, never calmly straightening a tie. Wealth is shown by pressing a black card or cash into someone's hand, never by the line \enquote{money is no object}.

\textbf{Humiliation and revenge props need high contrast.} Avoid neutral objects. Draw a red lipstick cross on a mirror or pull an apron over a suit. If a single frame cannot convey the emotion at a glance, change the prop. The formula is extreme color, cheap everyday object, and unusual use.

\textbf{Dropped props move the plot.} A flashlight falls and pushes one character into another's arms. A lighter is picked up and points to a suspect. If a prop cannot trigger a collision, closeness, or a misunderstanding, change it.

\textit{[Remaining rules omitted.]}
\end{rulecard}

\begin{rulecard}{Genre conventions \quad \texttt{farm-rural-romance.md}}{description 152 tokens, rules 994 tokens}

\textit{Description.} Read when the story is set on a farm, in the countryside, or on an agricultural estate. You must read it when writing conflict actions, eviction scenes, and choices of props or weapons in farm or rural scenes, because a rural setting must draw on what is at hand rather than generic middle-class props. It covers prop choice, romantic scene design, conflict props, and proposal scenes. It does not apply to romances set in cities, wealthy families, schools, or other non-rural settings.

The thrill of farm stories comes from spectacle created by extremely down-to-earth props, not from generic romantic material.

\textbf{Props must be rustic, never generic romance.} Rose petals, red wine, and candles are wrong. Take props from the farm itself, such as a shovel, a plow, a hay bale, a pumpkin, a rusty chain, or a bucket. The more rustic the prop, the stronger the visual spectacle. The more generic, the flatter the scene.

\textbf{Conflict weapons come from what is at hand.} When the heroine wants to drive someone off or attack, she uses whatever is nearby, hitting with a shovel, blocking a door with a hay bale, or spraying someone with a hose. Using farm tools reads more strongly on screen than pushing and pulling.

\textbf{A confession or proposal needs rural spectacle, and the prop must be on the male lead himself.} When a billionaire proposes in the village, skip the helicopter and diamond ring, and skip scarecrows or words spelled out on an empty field. Have him wear a silly mascot costume, put on a pumpkin head, or kneel from a tractor. The winning formula is that this important man himself becomes the ridiculous prop. The comic contrast must land on him, not on the background.

\textit{[Remaining rules omitted.]}
\end{rulecard}
\clearpage
\flushbottom

\section{Training Dynamics Results}\label{sec:appendix:dynamics}

This appendix gives the setup and complete results behind \Cref{sec:exp:dynamics}.
The twenty-novel run takes 58 backward micro-steps per epoch, and each micro-step commits a new version of the skill library.
For \Cref{fig:training_dynamics}, we evaluate the empty library and the versions after micro-steps 5, 10, 20, 30, 40, 58, 87, 116, 145, and 174.
Each version writes the first 30 episodes of every held-out novel, so all versions are compared on the same 180 episodes.

To put counts, densities, and rates on one scale, we standardize each of the four directional episode-level structure metrics against the empty library.
For each metric, we subtract the empty library's mean from a version's mean and divide by the empty library's per-episode standard deviation.
The structure score averages these four values, so the empty library scores 0. Human scripts use the same empty-library mean and standard deviation.
The score does not rise monotonically. It dips at micro-step 40 and holds near 0.58 through the second epoch before rising to 0.75 in the third epoch.

The library grows mostly in the first epoch. By its end, the library already has about 22{,}500 tokens, compared with about 28{,}300 at the end of training, in line with the shift from net additions to rewrites in \Cref{fig:training_dynamics}b.
\Cref{tab:epoch_full} evaluates the three epoch-end versions on all 385 held-out episodes. The last of them is +Ours in \Cref{tab:main}.
After one epoch, all six directional metrics already improve significantly over Base.
The full test set also matches the trend that \Cref{fig:training_dynamics}a shows on 180 episodes. Hard-cut endings and prop activation dip in the second epoch, and the third epoch gives the best value on five of the six directional metrics.
Bold marks the highest $\uparrow$ or lowest $\downarrow$ value among configurations, and $\sim$ metrics have no bold winner.

\begin{table}[h]
\centering
\caption{Held-out evaluation of twenty-novel libraries after each epoch on all 385 episodes. Count and density rows report group means with same-episode human MAE as $d$. Rate rows report yes-rates with signed percentage-point gaps to human as $d$. In the value columns, $^{*}$ marks paired differences from Base whose 95\% bootstrap intervals exclude zero, using raw counts, densities, or binary verdicts. The $d$ columns are supplementary comparisons with human originals.}
\label{tab:epoch_full}
\small
\setlength{\tabcolsep}{2.5pt}
\resizebox{\linewidth}{!}{%
\begin{tabular}{lrrrr>{\columncolor{black!6}}rrrrr}
\toprule
Metric & Base & Epoch 1 & Epoch 2 & Epoch 3 & Human & $d_{\text{Base}}$ & $d_{\text{Ep1}}$ & $d_{\text{Ep2}}$ & $d_{\text{Ep3}}$ \\
\midrule
\rowcolor{tabrule}\multicolumn{10}{l}{\textit{Line-level style}} \\
Longest dialogue volley (turns/ep)~$\sim$ & 4.5 & 4.3 & 4.1$^{*}$ & 4.2$^{*}$ & 3.6 & 1.86 & 1.75 & 1.70 & 1.72 \\
Dialogue line length (words)~$\sim$ & 5.6 & 7.6$^{*}$ & 6.6$^{*}$ & 7.3$^{*}$ & 11.9 & 6.53 & 4.77 & 5.61 & 4.99 \\
Voice-over cues (per ep)~$\sim$ & 0.35 & 0.59$^{*}$ & 0.25 & 0.52$^{*}$ & 0.22 & 0.48 & 0.67 & 0.39 & 0.58 \\
Unfilmable action lines (per ep)~$\downarrow$ & 1.2 & 0.3$^{*}$ & 0.3$^{*}$ & \textbf{0.2}$^{*}$ & 0.4 & 1.09 & 0.49 & 0.50 & 0.46 \\
Stimulus words (per ep)~$\uparrow$ & 3.3 & 5.5$^{*}$ & 5.6$^{*}$ & \textbf{6.3}$^{*}$ & 9.2 & 6.54 & 4.97 & 4.95 & 5.01 \\
\midrule
\rowcolor{tabjudge}\multicolumn{10}{l}{\textit{Episode-level structure}} \\
Plot conflicts (per 1k words)~$\uparrow$ & 2.10 & 3.37$^{*}$ & 3.58$^{*}$ & \textbf{3.69}$^{*}$ & 3.47 & 2.17 & 1.97 & 2.04 & 2.06 \\
Hard-cut ending rate~$\uparrow$ & 33.8\% & 62.6\%$^{*}$ & 54.8\%$^{*}$ & \textbf{66.8\%}$^{*}$ & 62.1\% & $-28.3$ & $+0.5$ & $-7.3$ & $+4.7$ \\
Cold-open rate~$\uparrow$ & 51.7\% & 75.8\%$^{*}$ & \textbf{76.1\%}$^{*}$ & 74.0\%$^{*}$ & 59.0\% & $-7.3$ & $+16.9$ & $+17.1$ & $+15.1$ \\
Prop-activation rate~$\uparrow$ & 46.0\% & 57.0\%$^{*}$ & 51.9\%$^{*}$ & \textbf{58.2\%}$^{*}$ & 57.4\% & $-11.4$ & $-0.4$ & $-5.5$ & $+0.8$ \\
Ending-twist rate~$\sim$ & 33.0\% & 36.4\% & 35.8\% & 30.6\% & 31.4\% & $+1.6$ & $+4.9$ & $+4.4$ & $-0.8$ \\
\bottomrule
\end{tabular}}

\end{table}

\begin{figure}[h]
\centering
\includegraphics[width=\linewidth]{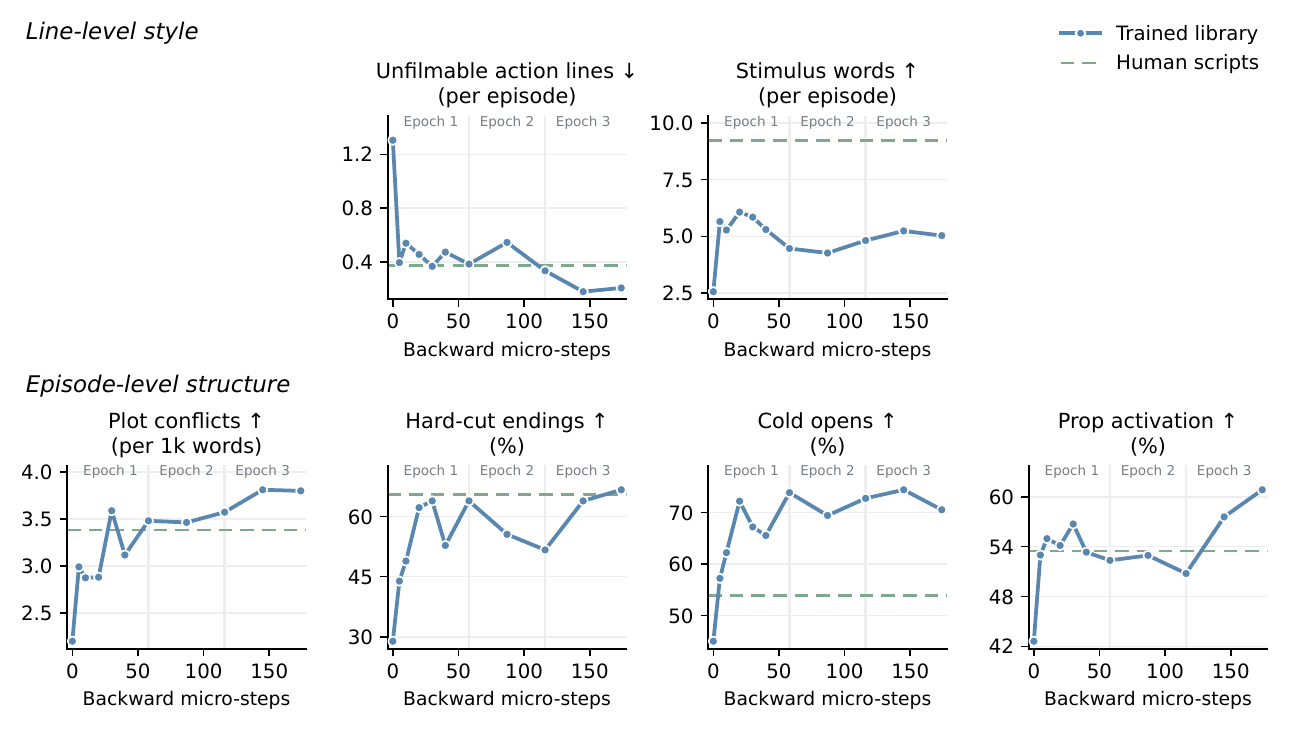}
\caption{Directional metrics of twenty-novel library snapshots across backward micro-steps, on the first 30 episodes of each held-out novel. Panels report mean counts or densities and binary yes-rates, and dashed lines mark human scripts on the same episodes.}
\label{fig:dynamics_metrics}
\end{figure}

\Cref{fig:dynamics_metrics} shows the four metrics behind the structure score, together with the two directional line-level style metrics that the score leaves out.
Within five micro-steps, unfilmable action lines fall from 1.31 to 0.39 per episode and stimulus words more than double, with smaller changes afterwards.

\clearpage

\section{Human Preference Evaluation}\label{sec:appendix:human}

We evaluate scripts generated using a Sonnet 4.6 skill library trained on twenty novels.
The study covers 45 distinct episodes from the six held-out novels, with 15 episodes for each comparison in \Cref{tab:human_counts}.
Five readers each assess all 45 pairs, yielding 225 judgments. Three are external short-drama practitioners (a screenwriter, an AI short-drama producer, and a TikTok editor), and two are paper authors serving as general readers.
\Cref{tab:human_per_reader} lists the votes of each reader.
Each pair presents two versions of the same episode with hidden group identities and randomized left--right placement. Readers select a preferred version or a tie and may provide a reason.
The three comparisons use different episodes, so differences between their win rates are descriptive rather than paired improvements on the same stories.

\begin{table}[h]
\centering
\caption{Human preference counts. Each comparison contains 15 episode pairs rated by five readers, giving 75 judgments per comparison.}
\label{tab:human_counts}
\small
\begin{tabular}{llrrr}
\toprule
Version A & Version B & A wins & Ties & B wins \\
\midrule
Empty library & Trained library & 17 & 9 & 49 \\
Empty library & Human & 9 & 9 & 57 \\
Trained library & Human & 21 & 7 & 47 \\
\bottomrule
\end{tabular}
\end{table}

\begin{table}[h]
\centering
\caption{Per-reader preference counts across the three comparisons. Each reader evaluates 15 pairs per comparison.}
\label{tab:human_per_reader}
\small
\setlength{\tabcolsep}{4pt}
\begin{tabular}{l rrr rrr rrr}
\toprule
& \multicolumn{3}{c}{Empty vs.\ Trained} & \multicolumn{3}{c}{Empty vs.\ Human} & \multicolumn{3}{c}{Trained vs.\ Human} \\
\cmidrule(lr){2-4} \cmidrule(lr){5-7} \cmidrule(lr){8-10}
Reader & E & Tie & T & E & Tie & H & T & Tie & H \\
\midrule
Screenwriter       & 2  & 3 & 10 & 1 & 4 & 10 & 9  & 2 & 4  \\
AI Producer         & 7  & 3 & 5  & 4 & 2 & 9  & 3  & 2 & 10 \\
TikTok Editor       & 5  & 1 & 9  & 1 & 1 & 13 & 4  & 2 & 9  \\
Author 1            & 0  & 2 & 13 & 2 & 2 & 11 & 3  & 1 & 11 \\
Author 2            & 3  & 0 & 12 & 1 & 0 & 14 & 2  & 0 & 13 \\
\midrule
Total               & 17 & 9 & 49 & 9 & 9 & 57 & 21 & 7 & 47 \\
\bottomrule
\end{tabular}
\end{table}

We score a Version~B win as $+1$, a tie as $0$, and a Version~A win as $-1$.
We average the scores within each episode, then average across episodes.
We obtain 95\% percentile intervals from 10{,}000 bootstrap resamples of the 15 episodes using seed 20260924, keeping all judgments for an episode together.
The intervals reflect variation across episodes for these fixed readers, not uncertainty over a population of new readers.

\begin{table}[h]
\centering
\caption{Net preference scores and 95\% bootstrap intervals. Positive scores favor Version~B.}
\label{tab:human_ci}
\small
\begin{tabular}{llrr}
\toprule
Version A & Version B & Net preference & 95\% interval \\
\midrule
Empty library & Trained library & $+42.7\%$ & $[+22.7\%, +61.3\%]$ \\
Empty library & Human & $+64.0\%$ & $[+42.7\%, +81.3\%]$ \\
Trained library & Human & $+34.7\%$ & $[+14.7\%, +53.3\%]$ \\
\bottomrule
\end{tabular}
\end{table}

\Cref{tab:human_ci} shows that all three intervals exclude zero. Readers prefer the trained library over the empty library, prefer human scripts over either generated version, and prefer human scripts over the trained library by a smaller margin than over the empty library.

\paragraph{Inter-rater agreement.} Each of the 45 script pairs is judged by five readers, which gives $\binom{5}{2}=10$ reader pairs per script pair and 450 in total. The pairwise agreement rate across these 450 reader pairs is 52\%, and Fleiss' $\kappa = 0.197$ with verdicts coded by screen position as left, tie, or right.
Readers differ in what they prioritize, yet four of the five prefer the trained version over the empty-library version.

\paragraph{Qualitative patterns from reader reasons.} Beyond the preference counts, we also examine why readers made their choices. Among the 225 judgments collected from the five readers, 142 include a written reason explaining their preference. We analyze these reasons and highlight two patterns: first, why the professional screenwriter sometimes prefers the trained version even over human originals, and second, where other practitioners feel the trained scripts still fall short of human writing.

The screenwriter is the most domain-relevant reader among the five. In the trained-library versus human comparison, she prefers the trained version in 9 of 15 pairs. Her reasons center on hooks, density, and characters. The human versions she rejects sometimes \enquote{cannot hold the ending hook}, have \enquote{low information density}, or let a flashback \enquote{interrupt the emotional build-up}. The trained versions she prefers \enquote{hold the hook}, have \enquote{higher plot density} and \enquote{more tension}, and give the characters \enquote{more striking personas}. These concerns overlap with what the proxy metrics measure, such as hard-cut endings and plot-conflict density.

On the other hand, when readers prefer the human version over the trained version, the most frequently cited issue is dialogue quality. Trained scripts sound \enquote{too literary}, \enquote{not like real speech}, or \enquote{like reading aloud}. Readers also note that human scripts occasionally deliver stronger emotional intensity in key moments. The skill library improves structural discipline, but dialogue naturalness and emotional expressiveness remain areas where there is still room for improvement.

\end{document}